\documentclass[10pt,twocolumn,letterpaper]{article}

\usepackage{wacv}              % To produce the CAMERA-READY version

\usepackage{algorithm}
\usepackage{algorithmic}

\usepackage{times}
\usepackage{epsfig}
\usepackage{amssymb}
\usepackage{booktabs}
\usepackage{tabularx}
\usepackage{multirow}
\usepackage{subcaption}
\usepackage[export]{adjustbox}
\usepackage{float}
\usepackage{dashrule}
\usepackage{arydshln}
\usepackage[utf8x]{inputenc}
\usepackage{anyfontsize}
\usepackage{t1enc}
\usepackage{mathtools, nccmath}

\usepackage{amsmath,amsfonts}
\usepackage{algorithmic}
\usepackage{algorithm}
\usepackage{array}
\usepackage{textcomp}
\usepackage{url}
\usepackage{verbatim}
\usepackage{graphicx}
\usepackage{animate}
\usepackage{bbding}
\usepackage{tablefootnote}

\usepackage{cuted}
\usepackage{capt-of}
\usepackage{caption}
\usepackage[table]{xcolor}

\usepackage{newfloat}
\usepackage{listings}

\newlength \mywidth
\newlength \myw
\definecolor{myblue}{HTML}{006EAF}
\definecolor{myred}{HTML}{CC0000}

\newcommand\na{\mbox{\scriptsize\color{gray} N/A}}

\newcommand{\ours}{DIVAD}

\definecolor{wacvblue}{rgb}{0.21,0.49,0.74}
\usepackage[pagebackref,breaklinks,colorlinks,allcolors=wacvblue]{hyperref}

\def\wacvPaperID{2956} % *** Enter the WACV Paper ID here
\def\confName{WACV}
\def\confYear{2026}

\title{Training Free Zero-Shot Visual Anomaly Localization via Diffusion Inversion}

\author{Samet Hicsonmez, Abd El Rahman Shabayek, Djamila Aouada \\
University of Luxembourg, Luxembourg\\
{\tt\small \{samet.hicsonmez,abdelrahman.shabayek,djamila.aouada\}@uni.lu}
}

\begin{document}
\maketitle
\begin{abstract}

Zero-Shot image Anomaly Detection (ZSAD) aims to detect and localise anomalies without access to any normal training samples of the target data. While recent ZSAD approaches leverage additional modalities such as language to generate fine-grained prompts for localisation, vision-only methods remain limited to image-level classification, lacking spatial precision. In this work, we introduce a simple yet effective training-free vision-only ZSAD framework that circumvents the need for fine-grained prompts by leveraging the inversion of a pretrained Denoising Diffusion Implicit Model (DDIM). Specifically, given an input image and a generic text description (e.g., "an image of an [object class]"), we invert the image to obtain latent representations and initiate the denoising process from a fixed intermediate timestep to reconstruct the image. Since the underlying diffusion model is trained solely on normal data, this process yields a normal-looking reconstruction. The discrepancy between the input image and the reconstructed one highlights potential anomalies. Our method achieves state-of-the-art performance on VISA dataset, demonstrating strong localisation capabilities without auxiliary modalities and facilitating a shift away from prompt dependence for zero-shot anomaly detection research. Code is available at~\url{https://github.com/giddyyupp/DIVAD}.
\end{abstract}    
\section{Introduction}
\label{sec:intro}

Image Anomaly Detection (AD) focusses on the identification and localisation of abnormal regions within images. 
It plays a critical role in numerous computer vision applications. 
For instance, in quality inspection for manufacturing, early detection of defects ensures product integrity and helps reduce operational costs~\cite{bergmann2018improving,visa,wang2024real}. 
Current unsupervised AD methods train models using all the available normal images in a dataset. However, collecting large numbers of normal data still requires careful inspection. Moreover, for each new object class, the data collection and model training should be conducted again. In order to reduce the data collection need, few-shot AD methods~\cite{li2024promptad_cvpr,winclip} are proposed where the methods are trained using even as few as a single normal image, but they still train a separate model for each new class.

\begin{figure}[t]
\centering
\includegraphics[width=1.0\linewidth]{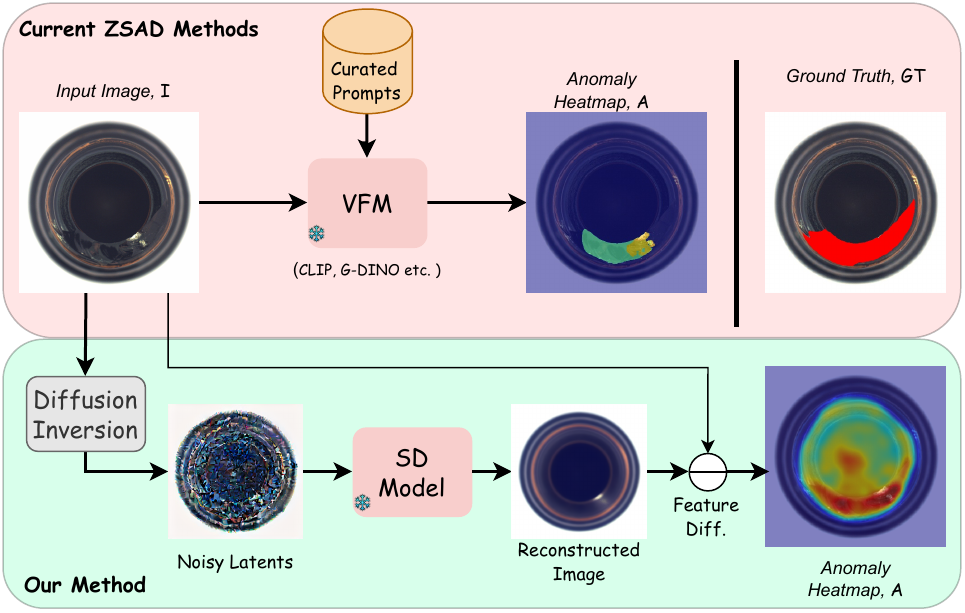}
\caption{\textbf{An overview of the current training-free ZSAD methods vs {\ours}}. Current methods utilize pretrained large Vision Foundation Models (VFM) with specially curated object specific \textit{guided-prompts}. Our method is free from any prompt tuning, and makes use of pretrained Stable Diffusion (SD) model to first invert a given image, and then reconstruct it through denoising.} 
\label{fig:teaser}
\end{figure}

Zero-shot AD (ZSAD) takes one step further and does not require any data collected from the target dataset. Current ZSAD methods~\cite{chen2023april,winclip,zhouanomalyclip,cao2025adaclip,li2024promptad,cao2025personalizing} could be grouped into two categories: a) cross-data training based, b) training-free. The former category utilises a fully annotated dataset (e.g., MVTec-AD~\cite{bergmann2019mvtec}) to train a model and test the trained model on the target dataset (e.g., VISA~\cite{visa}). This approach assumes the availability of a closely related dataset with extensive annotations. Furthermore, these training-based methods utilise additional modalities such as text and generally require extensive domain knowledge to generate the initial set of guided prompts to define both normality and abnormality. Most of the time, the guided prompts are curated using the already labelled target test sets. Similarly, except for a few works, e.g.,~\cite{maeday}, training-free ZSAD methods heavily rely on extensive domain knowledge to define, especially, the abnormality to extract useful textual features. For instance, WinCLIP~\cite{winclip} relies on over 1,000 handcrafted text prompts to achieve meaningful results, while AnomalyVLM~\cite{cao2025personalizing} directly utilises label information from the target test set to construct anomaly prompts.

As a prominent approach for AD, reconstruction-based methods~\cite{akcay2019ganomaly,zhang2023diffusionad,he2024diffusion,Fucka_2024_ECCV} utilise Generative Adversarial Networks (GANs)~\cite{gan} or diffusion models~\cite{ddpm,ddim}. Although these methods achieve very competitive performance, 1) they require large training sets, and 2) they train the models for long schedules, e.g., 1000 epochs for DiaD~\cite{he2024diffusion}. Another recent method, MAEDAY~\cite{maeday}, proposes using a pretrained Masked Auto-encoder to directly reconstruct a given image. However, their performance lags far behind that of current training-free ZSAD methods. Similar to MAEDAY, in our work, we directly use a pretrained diffusion model to reconstruct the given images for ZSAD with a novel formulation.

Our work is based on exploiting image inversion techniques. Inversion~\cite{ddim,dhariwal2021diffusion,mokady2023null,hertz2023prompt} of an image using diffusion models means finding an initial latent noise vector that produces the input image. Specifically, Denoising Diffusion Implicit Models (DDIM) inversion~\cite{ddim} uses DDIM sampling in the reverse order, i.e., adding noise to the input image. In order to find the corresponding noisy latent vector, the inversion should be run with a large number of time steps, such as $1000$. However, for efficiency, inversion is run for a few timesteps (e.g., 50), and sampling is done starting from an intermediate step (e.g., $20$) to end up with almost the same image as the input. Usually, the inversion process requires a simple text prompt describing the input image.   

In this work, we propose a simple training-free method utilising Diffusion Inversion for Visual Anomaly Detection, dubbed as {\ours}.
It does not rely on any prompt generation or guidance and utilises the diffusion inversion mechanism to reconstruct a given input image. As the diffusion model is trained to generate normal looking samples, the inversion process generates the \textit{normal version} of the input. The difference between the input and the inverted image denotes the locations of anomalies.

The proposed {\ours} has the following key contributions:
\begin{itemize}
    \item We propose a training-free ZSAD method that utilises diffusion inversion at its core.
    \item Our method does not require any domain knowledge to design special guided prompts, nor require any anomaly definitions.
    \item {\ours} achieves state-of-the-art results on the VISA and MPDD datasets and on-par performance on MVTec-AD compared to current training-free ZSAD methods.
\end{itemize}

\section{Related Work}
\label{sec:rel_work}

\subsection{Visual AD}
Existing visual AD methods generally fall into three categories: embedding-based, augmentation-based, and reconstruction-based approaches.

\noindent\textbf{Embedding-based methods} extract image or patch features using pretrained networks and then compare these features between normal and anomalous samples to identify anomaly pixels. Early work~\cite{s18010209} clustered features from ResNet~\cite{resnet} pretrained on ImageNet~\cite{deng2009imagenet}. Subsequent methods~\cite{spade,padim,roth2022towards,lee2022cfa} introduced memory banks to provide richer feature contexts. Flow-based approaches~\cite{gudovskiy2022cflow,lei2023pyramidflow} model the distribution of normal patch features or minimise distances between positive samples. Knowledge distillation is also used, where a student network learns to replicate a teacher model trained on normal data~\cite{Bergmann_Fauser_Sattlegger_Steger_2020,Wang_Wu_Cui_Shen_2021}.

\noindent\textbf{Augmentation-based methods} simulate anomalies by applying corruptions or noise to normal images~\cite{zavrtanik2021draem,zhang2024realnet} or their extracted features~\cite{liu2023simplenet,FAUAD}. For example, DRAEM~\cite{zavrtanik2021draem} uses Perlin noise~\cite{perlin} to generate synthetic anomalies and trains an autoencoder to reconstruct both clean and corrupted images, while a discriminator highlights abnormal regions.

\noindent\textbf{Reconstruction-based methods} learn to generate normal versions of inputs using models such as VAEs~\cite{bergmann2018improving}, GANs~\cite{akcay2019ganomaly,schlegl2019f}, vision transformers~\cite{zhang2023exploring}, and, more recently, diffusion models~\cite{wolleb2022diffusion,wyatt2022anoddpm,zhang2023diffusionad,he2024diffusion,hicsonmez2025vlmdiff}. During inference, these methods rely on the assumption that anomalies will lead to higher reconstruction errors. For example, Bergmann et al.~\cite{bergmann2018improving} showed that VAEs trained with an SSIM loss can detect pixel-level defects, while GANomaly~\cite{akcay2019ganomaly} uses a GAN to detect image-level anomalies. ViTAD~\cite{zhang2023exploring} integrates vision transformers into a VAE framework and achieves state-of-the-art results.

All these methods require a large labelled dataset, and most of them train for very long epochs to achieve satisfactory results. 

\subsection{Zero-Shot visual AD} ZSAD methods use extensive language guidance with the help of domain knowledge. They can be categorised into two groups: a) cross-data training-based methods and b) training-free methods.

\noindent\textbf{Cross-data training based methods} utilise an external fully labelled dataset with both normal and anomaly samples to train a model. These methods~\cite{chen2023april,zhouanomalyclip,gu2024filo,cao2025adaclip,li2024promptad} make use of pretrained large-scale Vision Language Models (VLMs)~\cite{clip}. They use frozen image encoders and make use of prompt learning strategies to inject additional tokens into text encoders to better represent normality and abnormality. AnomalyCLIP~\cite{zhouanomalyclip} introduces $12$ separate object-agnostic learnable text prompt templates to represent both normal and abnormal data. They train the model on the test sets of external datasets and evaluate it on the target test set. Similarly, FiLo~\cite{gu2024filo} creates learnable prompts but guides the training with a possible anomaly location extracted using Grounding DINO~\cite{liu2024grounding}. The detected anomaly location is fed to both the anomaly prompts in the form of bounding box representation and to the visual encoder to focus on this region only. AdaCLIP~\cite{cao2025adaclip} extends the idea of learnable text tokens to learnable image tokens. Moreover, in addition to learnable static tokens that are common to all the images, they use frozen CLIP as a Dynamic Prompt Generator to extract image-specific class tokens, which are incorporated into the learnable tokens.

\noindent\textbf{Training-free methods} do not need any external datasets to train a model, however, they still heavily rely on domain-specific knowledge to devise guided prompts to extract the anomaly regions. WinCLIP~\cite{winclip} creates around 1000 text prompts to cover every possible anomaly definition. Moreover, it uses label information (e.g., broken, contaminated, cracked) available in the dataset, which requires a domain expert to annotate, to generate the guided prompts. ALFA~\cite{zhu2024llms} uses a pretrained LLM to generate class-specific normal and abnormal prompts. They query a given test image to the LLM with the prompt, \textit{Describe what a normal/abnormal image of [cls] looks like?} to extract the definitions. These LLM prompts are then directly used without any learning in the WinCLIP-like pipeline.    
AnomalyVLM~\cite{cao2025personalizing} proposes a two-stage approach: first, the possible anomaly regions (bounding boxes) are extracted using Grounding DINO along with domain expert guided textual prompts, and then they use Segment Anything (SAM)~\cite{kirillov2023segment} to segment the anomalies in this region using heuristics. 

Different from previous methods, MAEDAY~\cite{maeday} does not rely on any textual input. They use a plain Masked Auto Encoder (MAE) trained on ImageNet~\cite{deng2009imagenet}. During inference, they mask a large portion of an input image and let MAE fill in the masked regions to generate a complete image. Finally, the difference between the input and the reconstructed image denotes the anomalous regions. 

Our method follows a similar approach and does not require any form of expert-generated text prompts. {\ours} reconstructs the input through DDIM inversion and calculates the pixel-wise feature differences between the input and generated images to segment the anomalous regions.

\subsection{Diffusion Inversion} Latent vector manipulation based image editing has been successfully applied in GAN-based methods through inversion~\cite{zhu2016generative,brock2017neural,abdal2019image2stylegan,abdal2020image2stylegan++,gu2020image,wu2021stylespace}. Recently, the same approach is employed for diffusion methods~\cite{dhariwal2021diffusion,mokady2023null,hertz2023prompt,couairon2023diffedit} and achieves remarkable performance. The goal of diffusion inversion is to manipulate a given image by changing desired parts of the input prompt, e.g., replacing a cat with a dog. The very first DDIM or DDPM-based inversion~\cite{dhariwal2021diffusion} falls short when a big guidance value is used as the predicted noise with the input prompt, and denoising with the edited prompt diverges based on the strength of the parameter. Null text inversion~\cite{mokady2023null} overcomes this issue by minimising the predicted noise with the input prompt and an empty prompt by optimising the empty text condition. Prompt-to-prompt~\cite{hertz2023prompt} decides which parts of the image should be edited based on the input and edited prompts by checking the attention maps. DiffEdit~\cite{couairon2023diffedit} takes a similar approach and generates a mask to decide where the editing should happen. SDEdit~\cite{mengsdedit} directly adds gaussian noise to the given stroke painting image, and then denoises it using a finetuned diffusion model to generate the corresponding color image. 

Our approach is similar to the SDEdit process, where we add sampled noise to the input image latent and denoise it. However, in our case, there is no user guidance in any form.

\section{Method}
\label{sec:method}

\subsection{Problem Definition}

For a given image $I\in \mathbb{R}^{H \times W \times C}$, Visual AD aims to perform 1) image-level classification and 2) pixel-level segmentation of anomalies. 
In this work, we focus on developing a  training-free ZSAD method which \textbf{does not have any learnable parameters and does not require any labelled images}.  
Moreover, the method should be category-agnostic, i.e. not requiring any adaptation based on the characteristics of a specific object type. 

\begin{figure*}[t]
\centering
\includegraphics[width=1.0\linewidth]{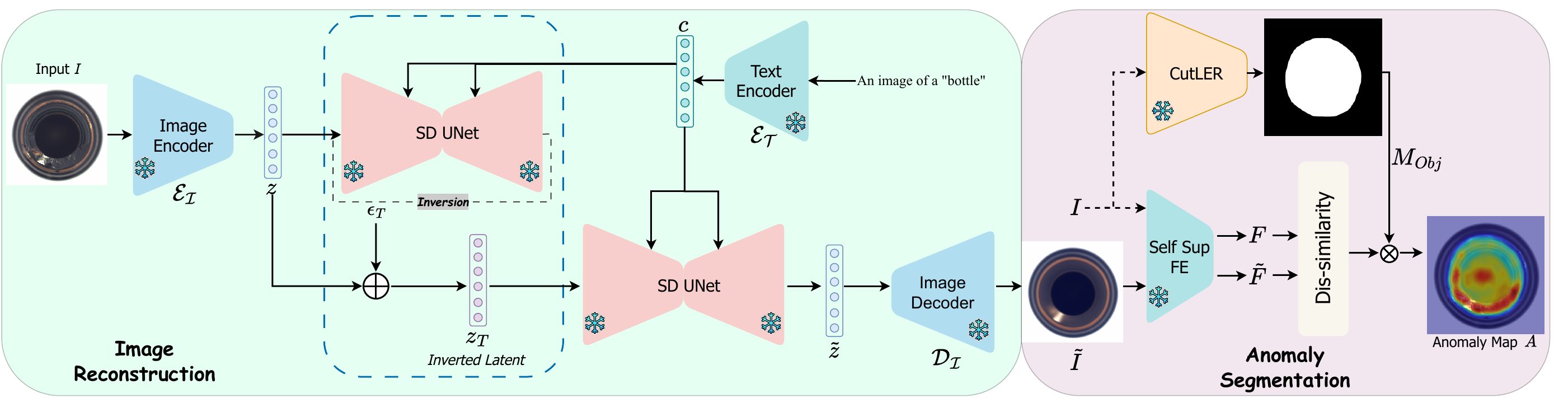}
\caption{\textbf{The processing pipeline of {\ours}}. Given test image first encoded to latent representation, the SD UNet~\cite{ldm} predicts the noise $\epsilon_T$ for $T$ timesteps using \textit{DDIM Inversion} process. Predicted noise is added to the input image latent $z$ to obtain inverted latent $z_T$. Then, this inverted latent is fed to the SD UNet model for denoising. Denoised latent is forwarded to the image decoder to recover a clean normal image. Instead of pixel wise comparison of input and reconstructed images which generates noisy regions, we perform segmentation on feature-level by utilizing a self-supervised model. Additionally, we make use of an object segmenter (CutLER~\cite{wang2023cut}) to improve the segmentation performance.} 
\label{fig:model}
\end{figure*}

\subsection{Preliminaries}

This section provides an overview of the core components and terminology related to text-guided diffusion models, including the denoising objective, classifier-free guidance, and DDIM inversion. These concepts form the foundation for our proposed approach and are essential for understanding the methodology presented in subsequent sections.

\paragraph{\textbf{Text-Guided Diffusion Models.}}
Text-conditioned diffusion models aim to synthesize images by progressively refining a noise vector \( z_t \) into an  output image \( z_0 \), guided by a textual prompt \( \mathcal{P} \). The denoising network \( \epsilon_\theta \) is trained to predict the added noise \( \epsilon\), following the objective:
\begin{equation}
\min_\theta \mathbb{E}_{z_0, \epsilon \sim \mathcal{N}(0,I), t \sim \text{Uniform}(1,T)} \left\| \epsilon - \epsilon_\theta(z_t, t, c) \right\|^2,
\end{equation}
where \( c = \phi(\mathcal{P}) \) is the embedding of the text condition, and \( z_t \) is a noisy version of a clean latent sample \( z_0 \), obtained by perturbing it with gaussian noise at timestep \( t \).
During inference, starting from pure noise \( z_T \), the model progressively denoises the latent through \( T \) iterative steps using the trained network.

To enable accurate reconstruction of real images, we adopt deterministic DDIM sampling~\cite{ddim}, which performs the following update:

\begin{equation}
z_{t-1} = \sqrt{\frac{\alpha_{t-1}}{\alpha_t}} z_t + \left( \sqrt{\frac{1}{\alpha_{t-1}} - 1} - \sqrt{\frac{1}{\alpha_t} - 1} \right) \cdot \epsilon_\theta(z_t, t, c).
\end{equation}
Here, \( \alpha_t \) represents the variance schedule. 

Unlike traditional models operating in pixel space, we leverage the Stable Diffusion (SD) model~\cite{ldm}, which conducts diffusion in a learned latent space. An image \( x_0 \) is first encoded into a latent vector \( z = \mathcal{E}_I(x_0) \), processed via the diffusion model (i.e. noise addition and denoising using UNet), and then decoded back using \( x_0 = \mathcal{D}_I(z) \).

\paragraph{\textbf{Classifier-Free Guidance.}}
To enhance the effect of the text conditioning during generation, Ho et al.~\cite{ho2021classifier} proposed classifier-free guidance. This approach interpolates between conditional (i.e. using a text prompt) and unconditional (i.e. using an empty text) denoising predictions. Let \( \varnothing = \phi(``") \) denote the embedding of an empty prompt and \( w \) be the guidance scale, then:
\begin{equation}
\tilde{\epsilon}_\theta(z_t, t, c, \varnothing) = w \cdot \epsilon_\theta(z_t, t, c) + (1 - w) \cdot \epsilon_\theta(z_t, t, \varnothing).
\end{equation}
In practice, a typical value such as \( w=3.5\) is used for text-to-image generation in SD, which we also use in our experiments.

\paragraph{\textbf{DDIM Inversion.}}
To reconstruct a known image from its latent encoding, we employ DDIM inversion~\cite{dhariwal2021diffusion,ddim}. Under the assumption that the underlying Ordinary Differential Equation process can be approximately reversed, the inversion proceeds as follows:
\begin{equation}
z_{t+1} = \sqrt{\frac{\alpha_{t+1}}{\alpha_t}} z_t + \left( \sqrt{\frac{1}{\alpha_{t+1}} - 1} - \sqrt{\frac{1}{\alpha_t} - 1} \right) \cdot \epsilon_\theta(z_t, t, c).
\end{equation}
This formulation mirrors the forward process in reverse, from a clean latent code \( z \) which is typically obtained by encoding a real image, toward a noisy latent \( z_T \). This enables direct inspection and manipulation of real data within the diffusion framework.

\renewcommand{\arraystretch}{1}

\begin{figure*}[t]
\captionsetup[subfigure]{labelformat=empty}
\centering
\resizebox{1.0\textwidth}{!}{
\begin{tabular}{l@{}c@{}c@{}c@{}c@{}c@{}c@{}c@{}c@{}c@{}c}
& \tiny{bottle} & \tiny{cable} & \tiny{carpet} & \tiny{grid} & \tiny{metal nut} & \tiny{screw} & \tiny{toothbrush} & \tiny{transistor}& \tiny{wood}  & \tiny{zipper} \\

{\rotatebox[origin=t]{90}{\textit{{\tiny I}}}}  & 
\includegraphics[width=\mywidth,  ,valign=m, keepaspectratio,] {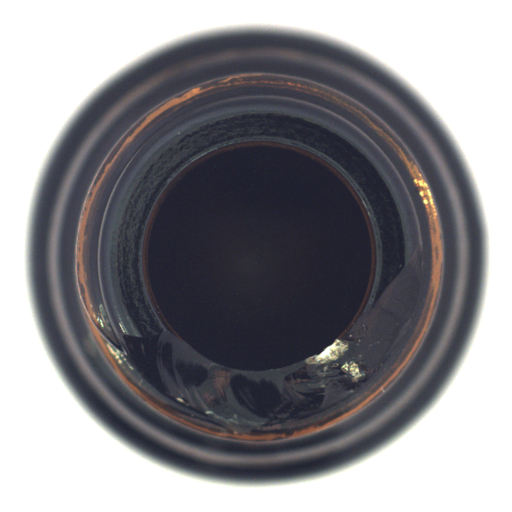} &
\includegraphics[width=\mywidth,  ,valign=m, keepaspectratio,] {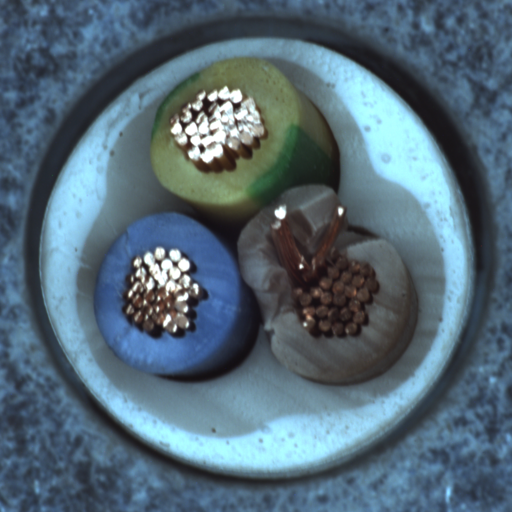} &
\includegraphics[width=\mywidth,  ,valign=m, keepaspectratio,] {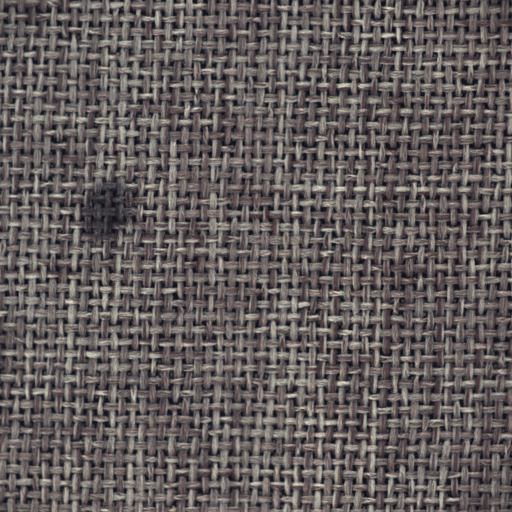} &
\includegraphics[width=\mywidth,  ,valign=m, keepaspectratio,] {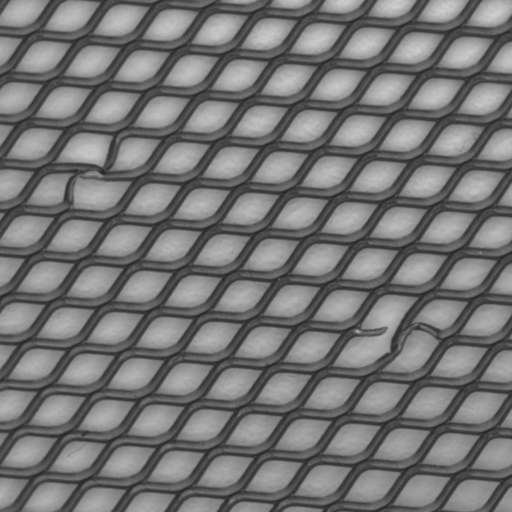} &
\includegraphics[width=\mywidth,  ,valign=m, keepaspectratio,] {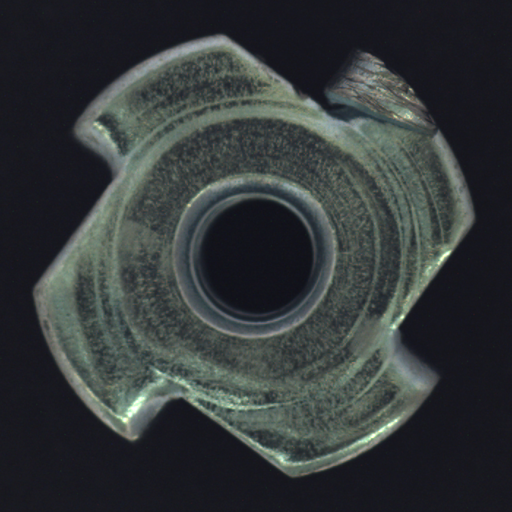} &
\includegraphics[width=\mywidth,  ,valign=m, keepaspectratio,] {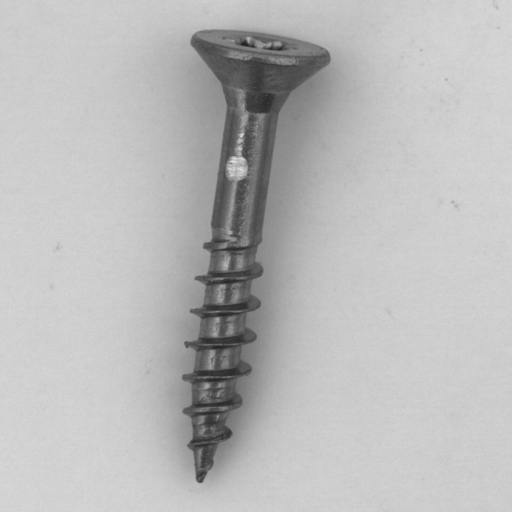} &
\includegraphics[width=\mywidth,  ,valign=m, keepaspectratio,] {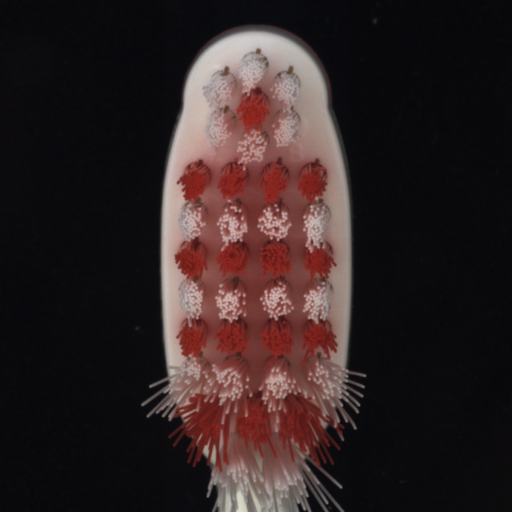} &
\includegraphics[width=\mywidth,  ,valign=m, keepaspectratio,] {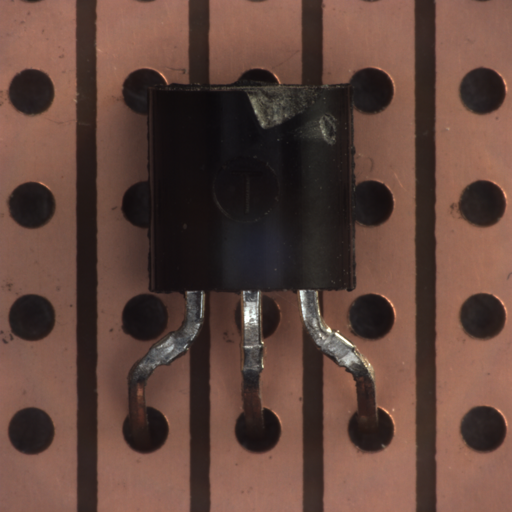} &
\includegraphics[width=\mywidth,  ,valign=m, keepaspectratio,] {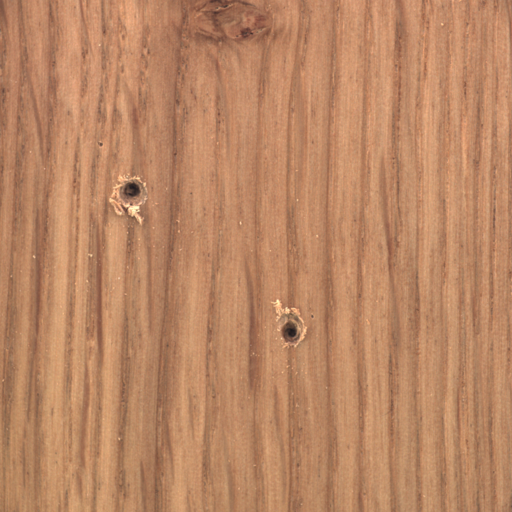} &
\includegraphics[width=\mywidth,  ,valign=m, keepaspectratio,] {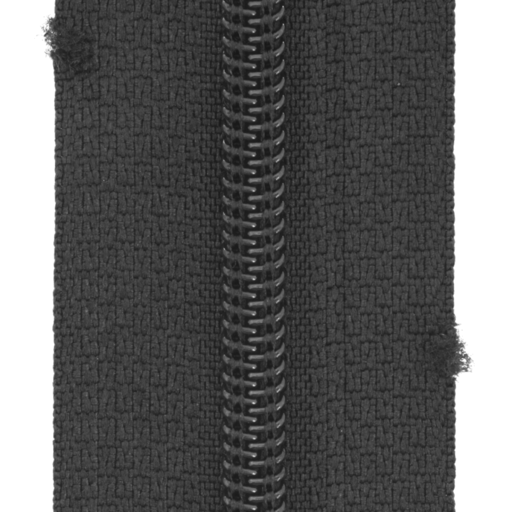} 
\\

{\rotatebox[origin=t]{90}{\textit{\textbf{ \tiny Inv.}}}}&  
\includegraphics[width=\mywidth,  ,valign=m, keepaspectratio,] {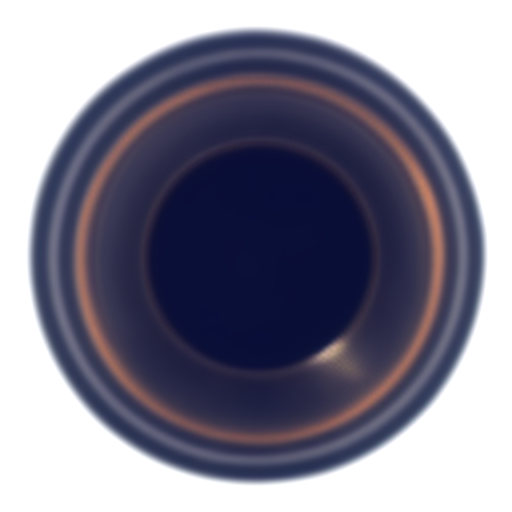} &
\includegraphics[width=\mywidth,  ,valign=m, keepaspectratio,] {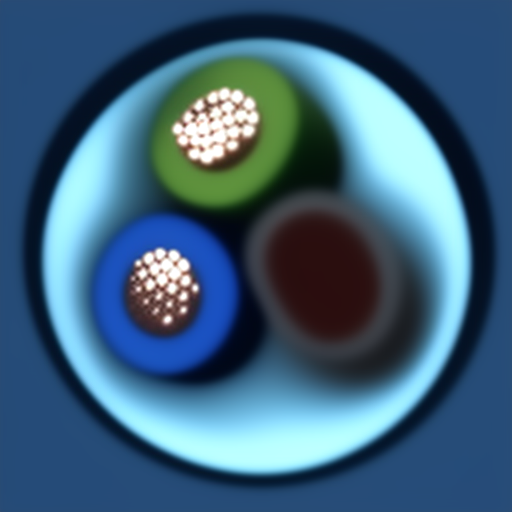} &
\includegraphics[width=\mywidth,  ,valign=m, keepaspectratio,] {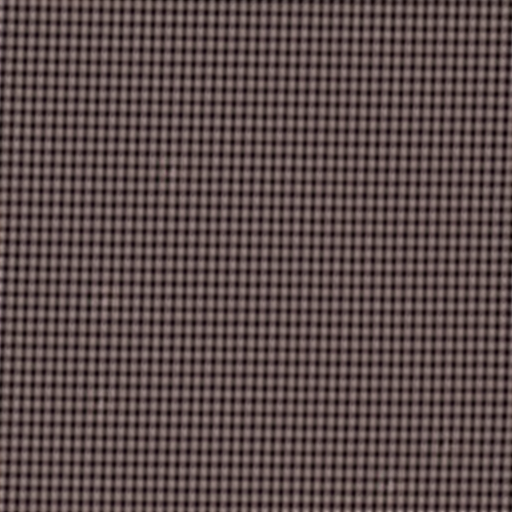} &
\includegraphics[width=\mywidth,  ,valign=m, keepaspectratio,] {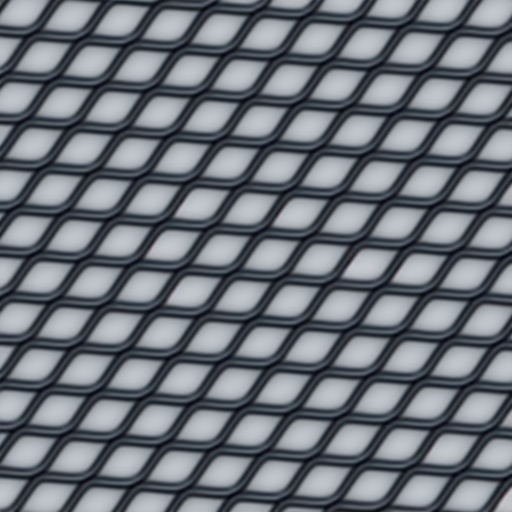} &
\includegraphics[width=\mywidth,  ,valign=m, keepaspectratio,] {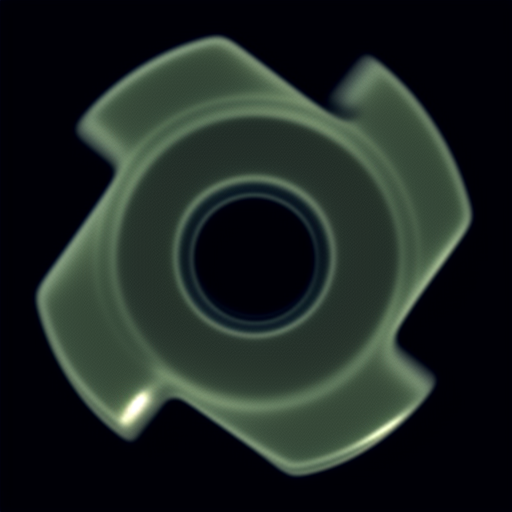} &
\includegraphics[width=\mywidth,  ,valign=m, keepaspectratio,] {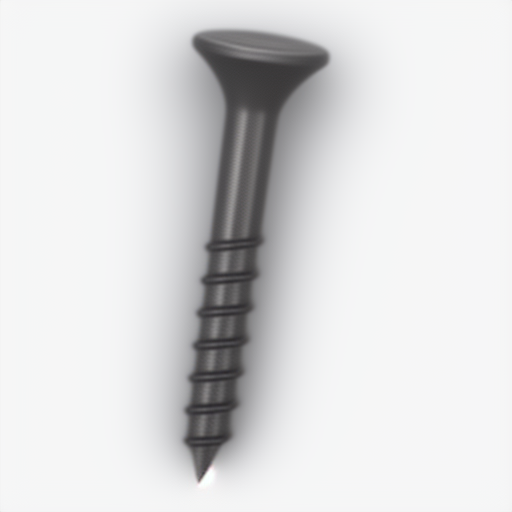} &
\includegraphics[width=\mywidth,  ,valign=m, keepaspectratio,] {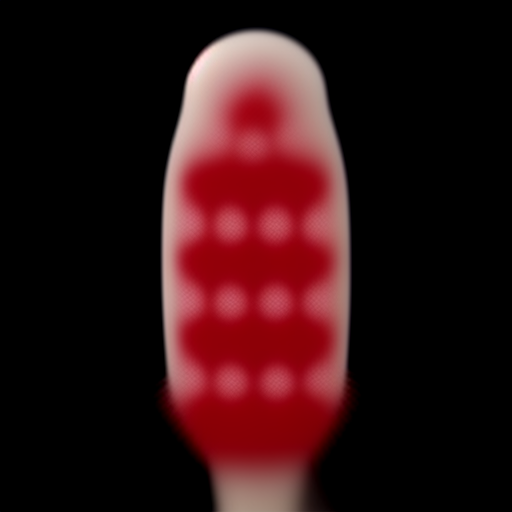} &
\includegraphics[width=\mywidth,  ,valign=m, keepaspectratio,] {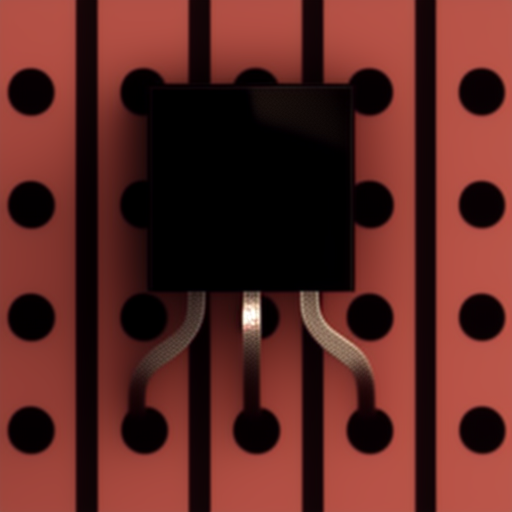} &
\includegraphics[width=\mywidth,  ,valign=m, keepaspectratio,] {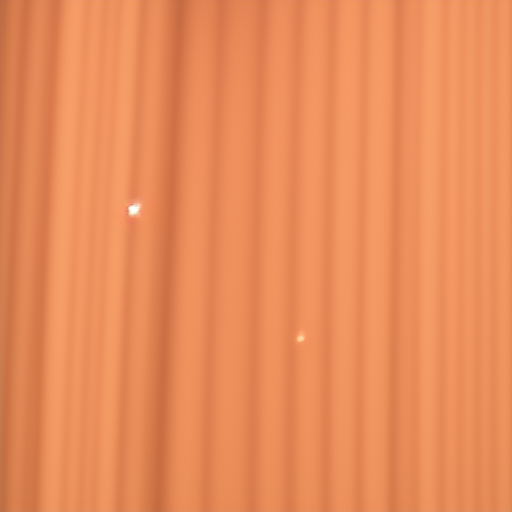} &
\includegraphics[width=\mywidth,  ,valign=m, keepaspectratio,] {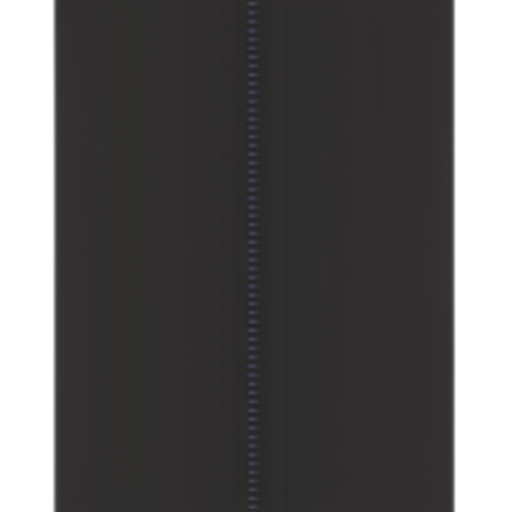} 
\\

{\rotatebox[origin=t]{90}{\textit{\textbf{\tiny {Pred.}}}}} & 
\includegraphics[width=\mywidth,  ,valign=m, keepaspectratio,] {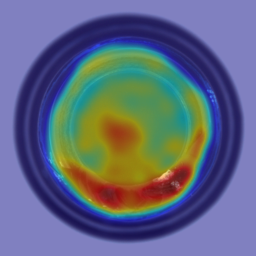} &
\includegraphics[width=\mywidth,  ,valign=m, keepaspectratio,] {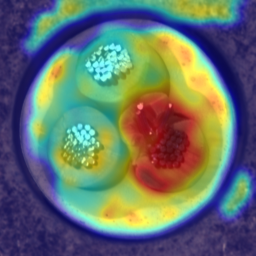} &
\includegraphics[width=\mywidth,  ,valign=m, keepaspectratio,] {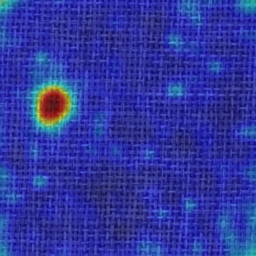} &
\includegraphics[width=\mywidth,  ,valign=m, keepaspectratio,] {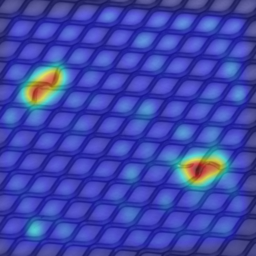} &
\includegraphics[width=\mywidth,  ,valign=m, keepaspectratio,] {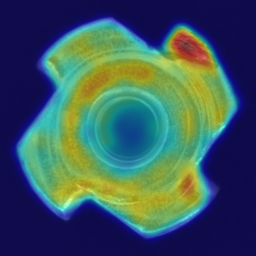} &
\includegraphics[width=\mywidth,  ,valign=m, keepaspectratio,] {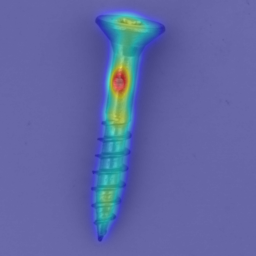} &
\includegraphics[width=\mywidth,  ,valign=m, keepaspectratio,] {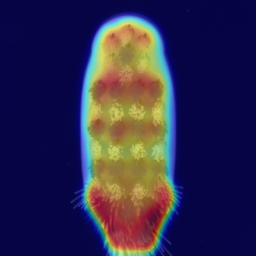} &
\includegraphics[width=\mywidth,  ,valign=m, keepaspectratio,] {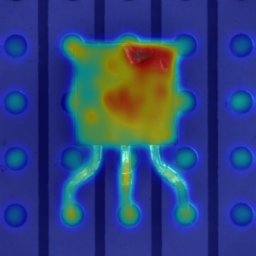} &
\includegraphics[width=\mywidth,  ,valign=m, keepaspectratio,] {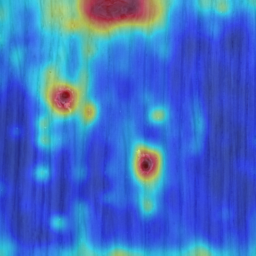} &
\includegraphics[width=\mywidth,  ,valign=m, keepaspectratio,] {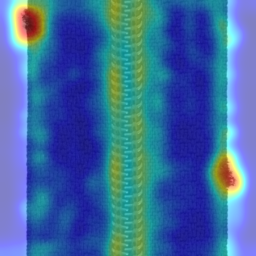} 
\\

{\rotatebox[origin=t]{90}{\textit{\textbf{\tiny GT}}}} & 
\includegraphics[width=\mywidth,  ,valign=m, keepaspectratio,] {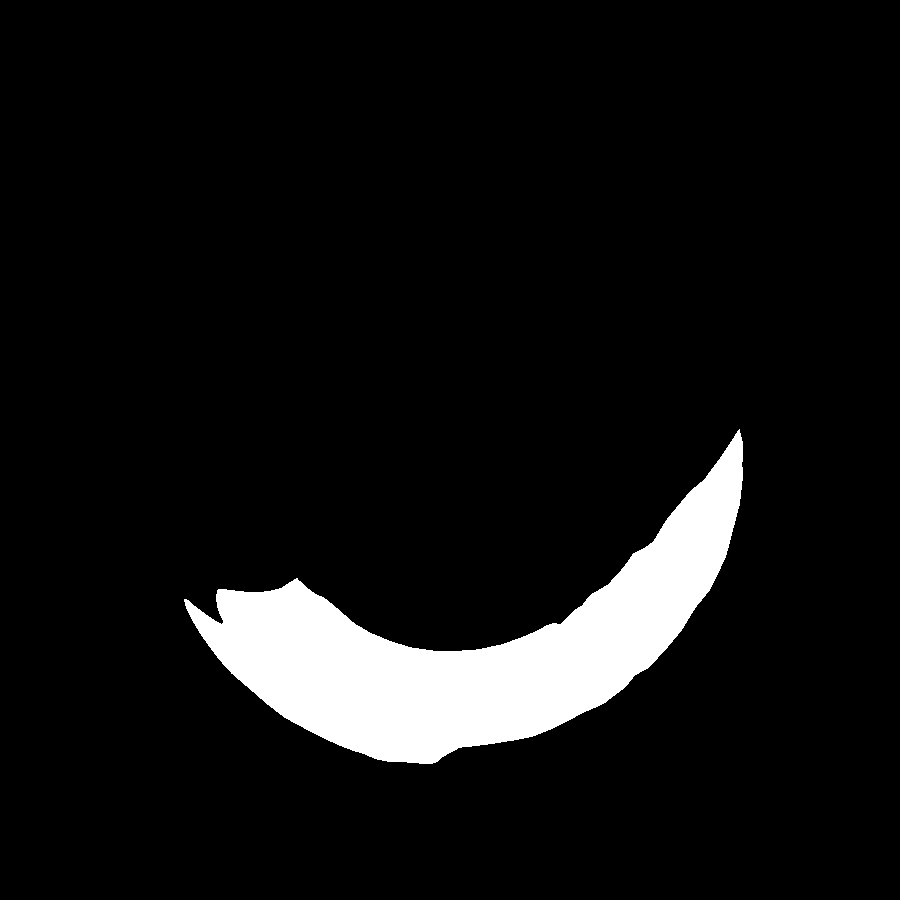} &
\includegraphics[width=\mywidth,  ,valign=m, keepaspectratio,] {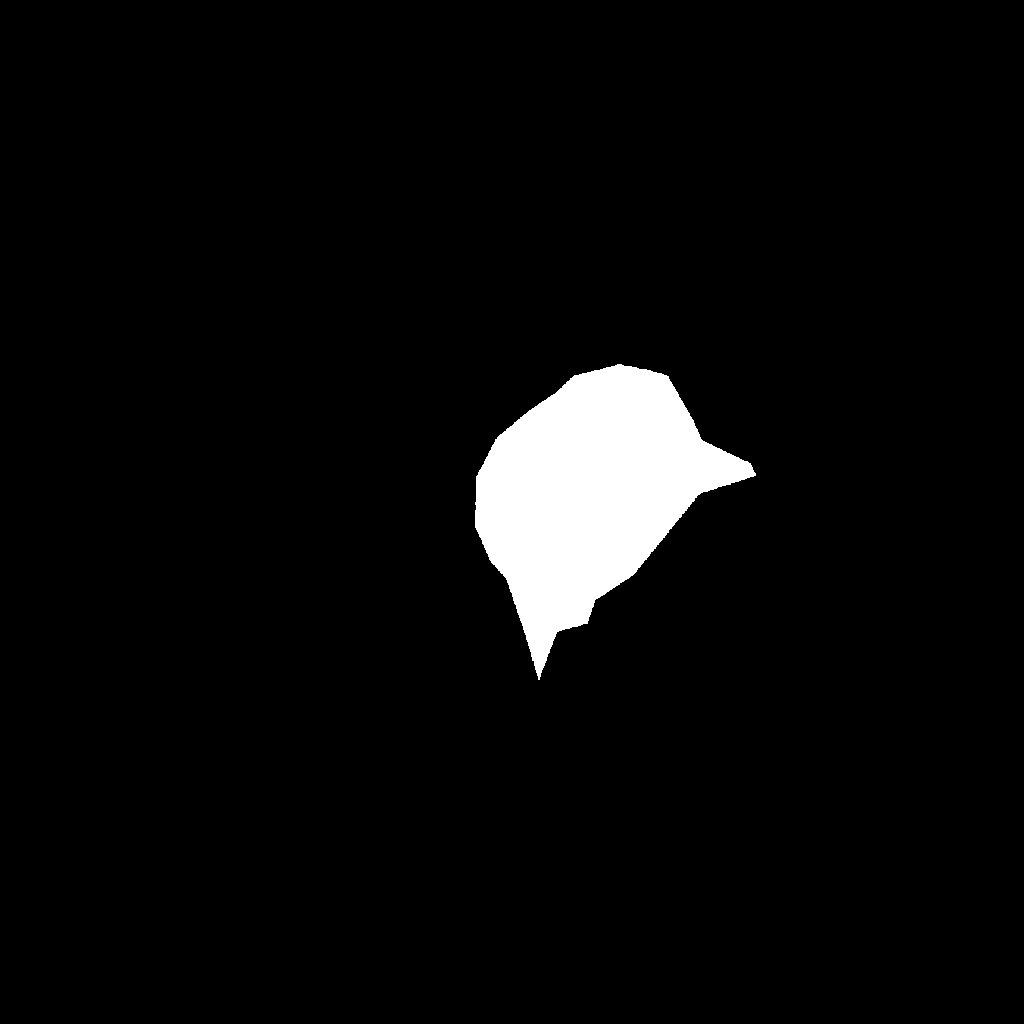} &
\includegraphics[width=\mywidth,  ,valign=m, keepaspectratio,] {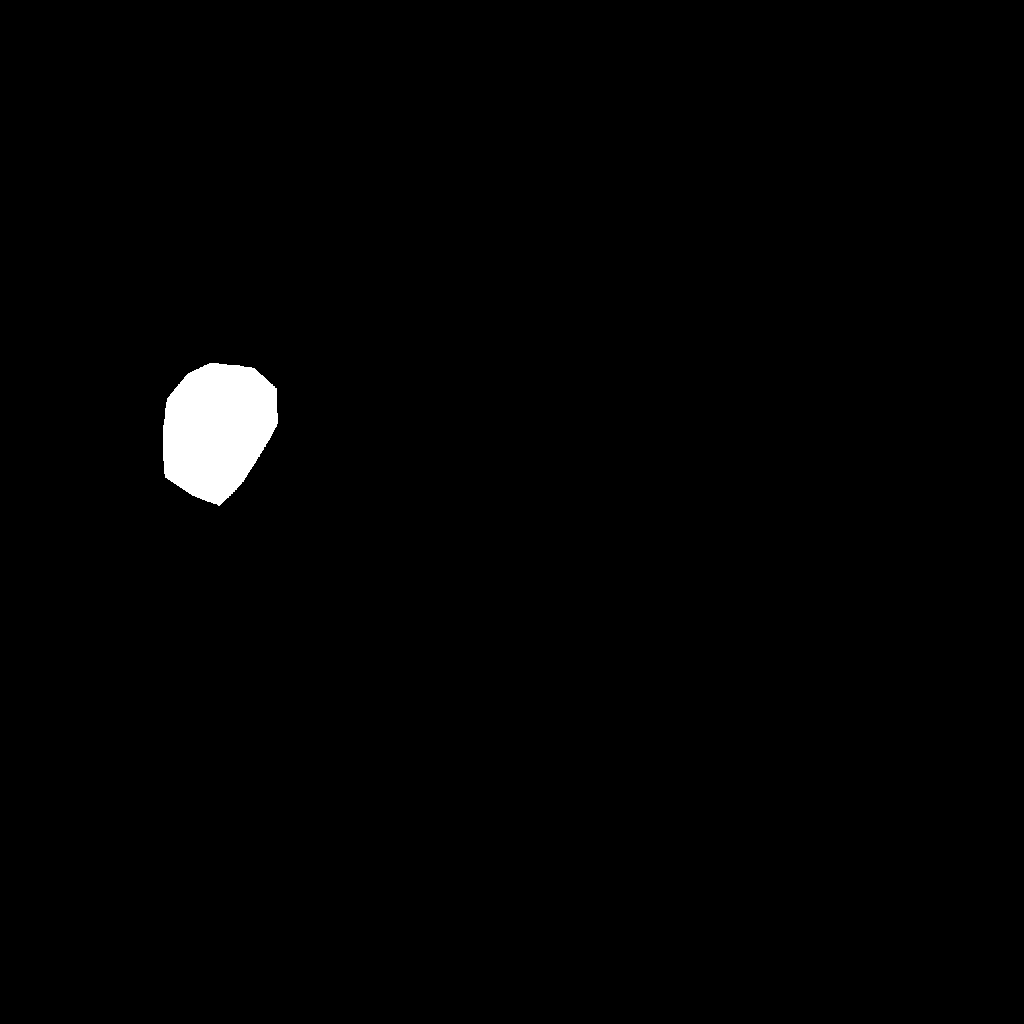} &
\includegraphics[width=\mywidth,  ,valign=m, keepaspectratio,] {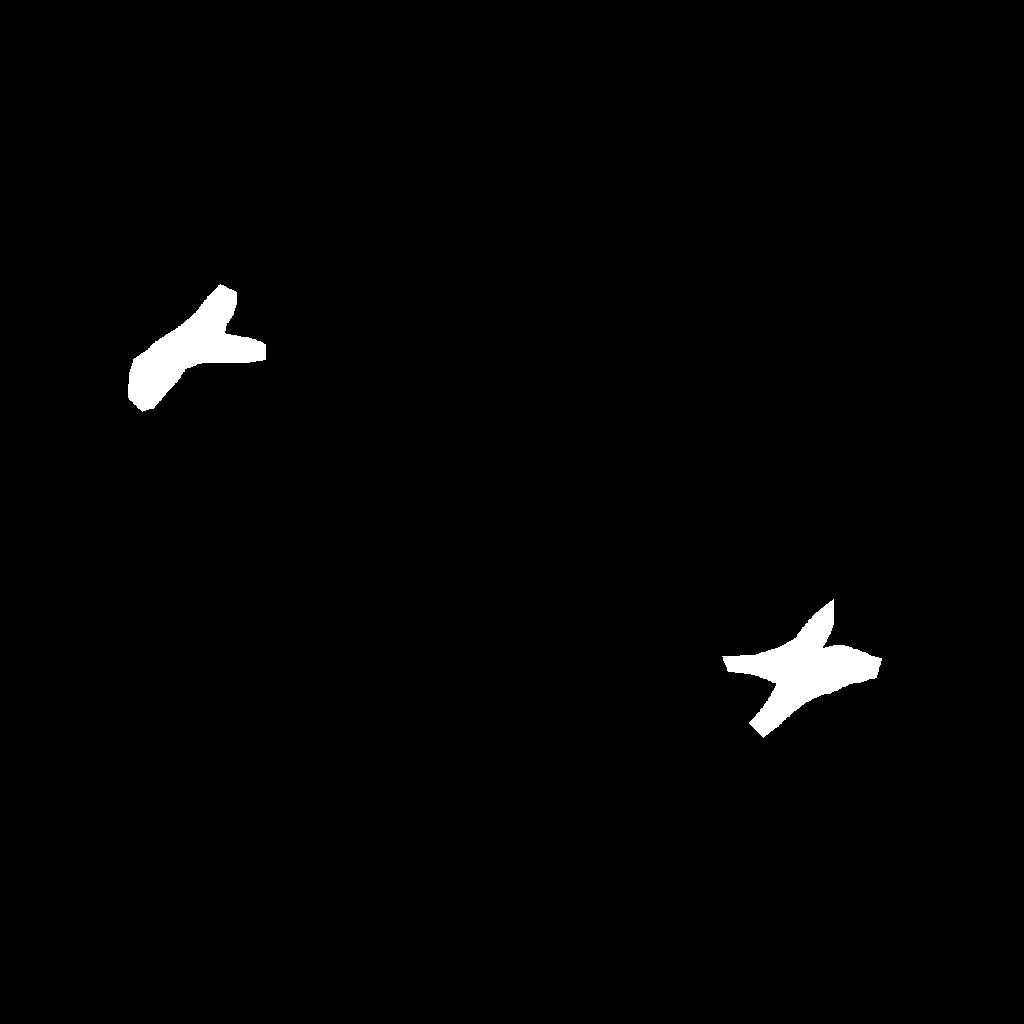} &
\includegraphics[width=\mywidth,  ,valign=m, keepaspectratio,] {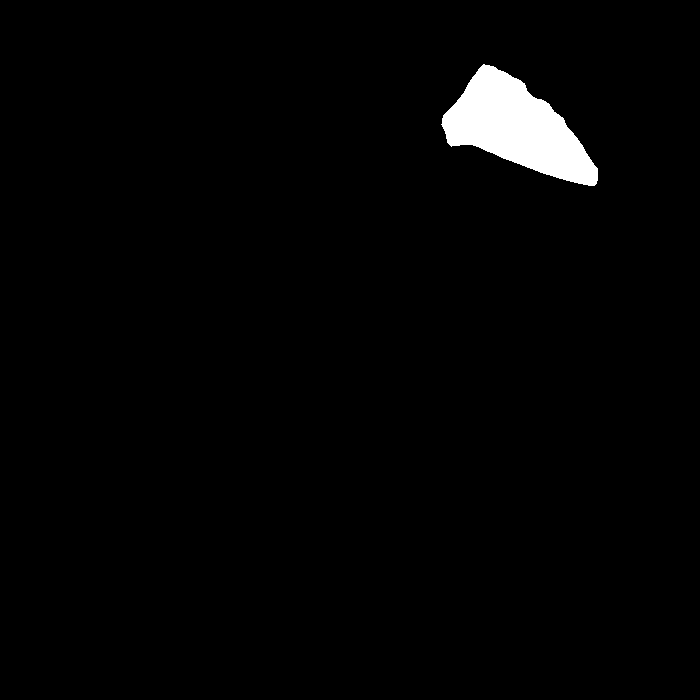} &
\includegraphics[width=\mywidth,  ,valign=m, keepaspectratio,] {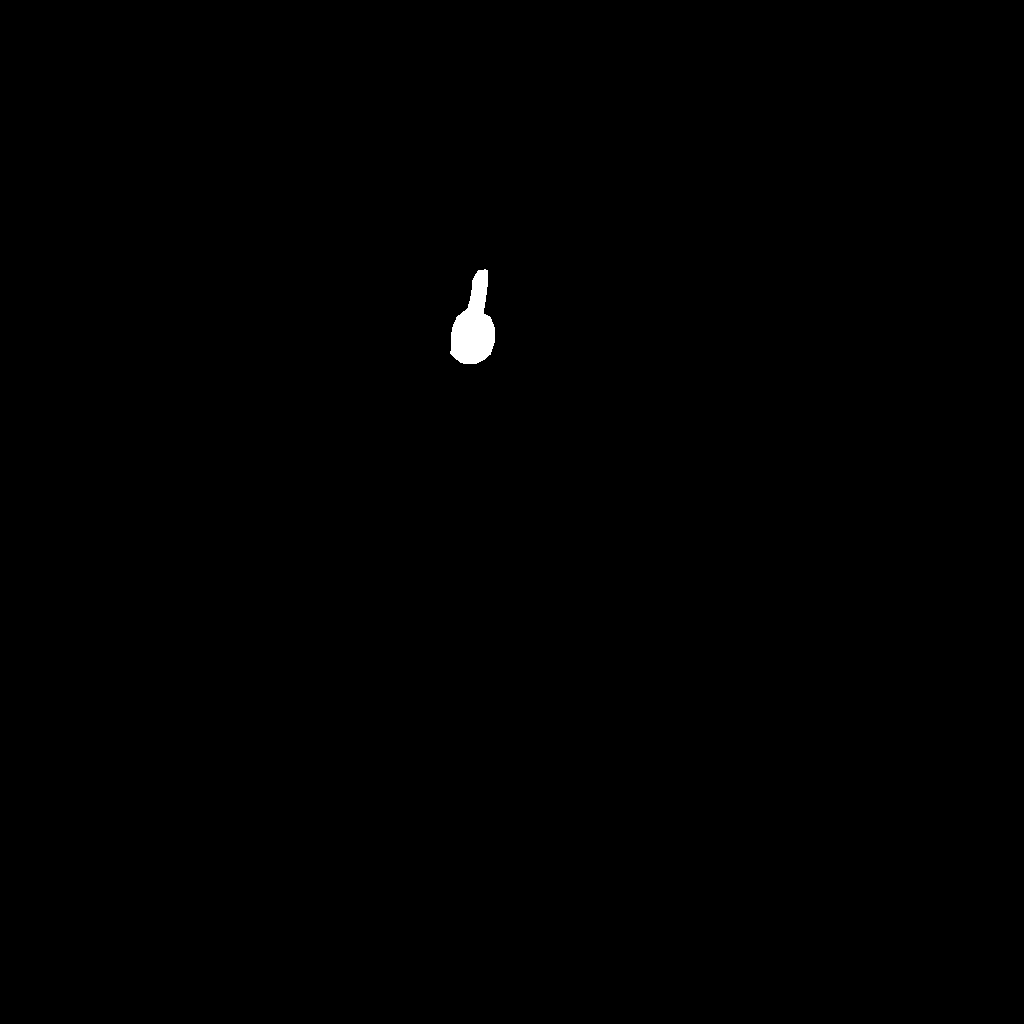} &
\includegraphics[width=\mywidth,  ,valign=m, keepaspectratio,] {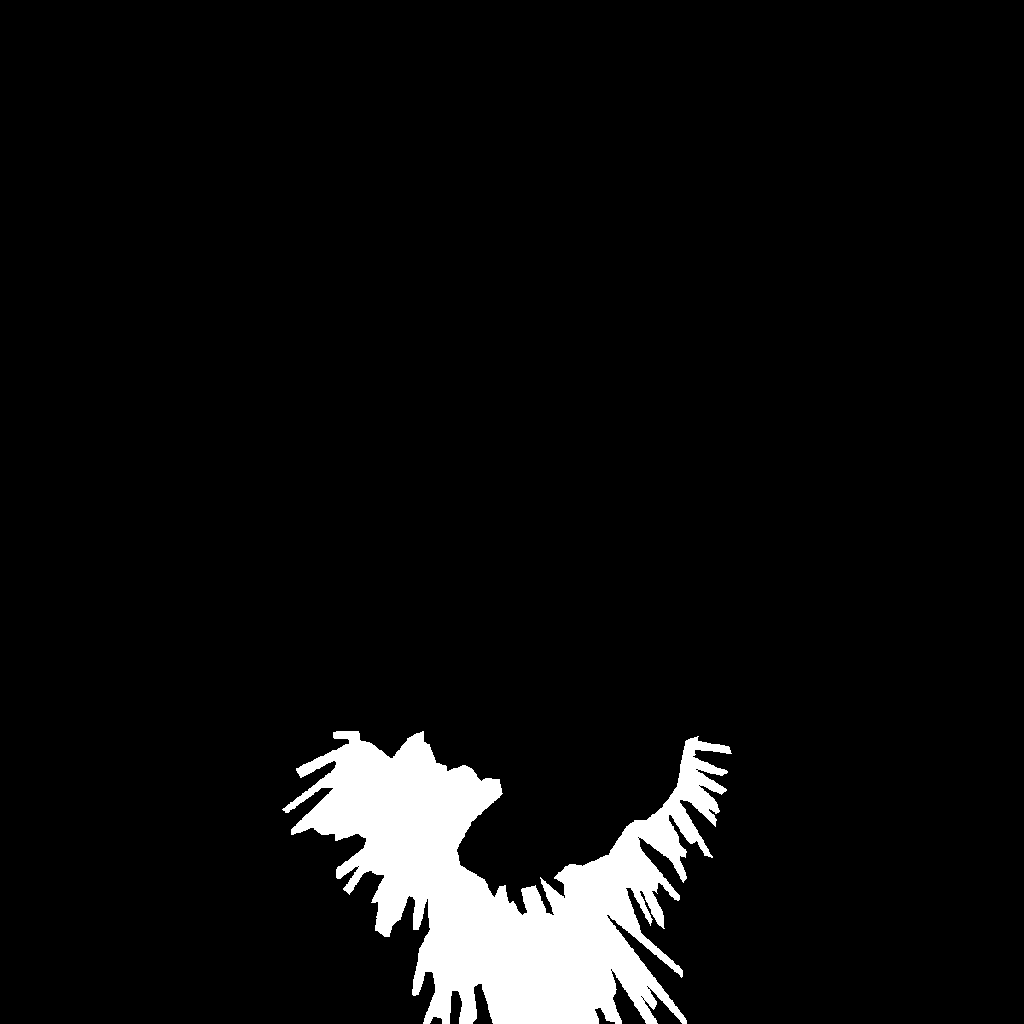} &
\includegraphics[width=\mywidth,  ,valign=m, keepaspectratio,] {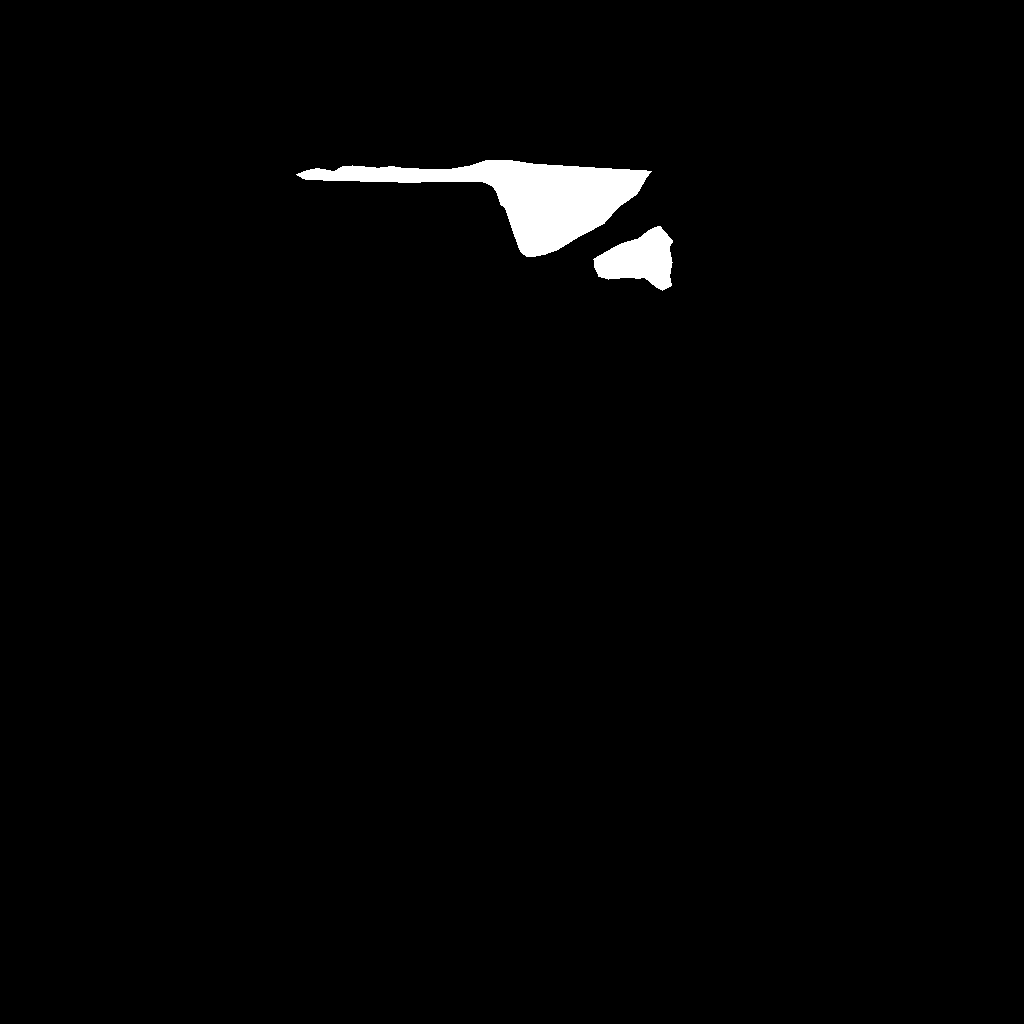} &
\includegraphics[width=\mywidth,  ,valign=m, keepaspectratio,] {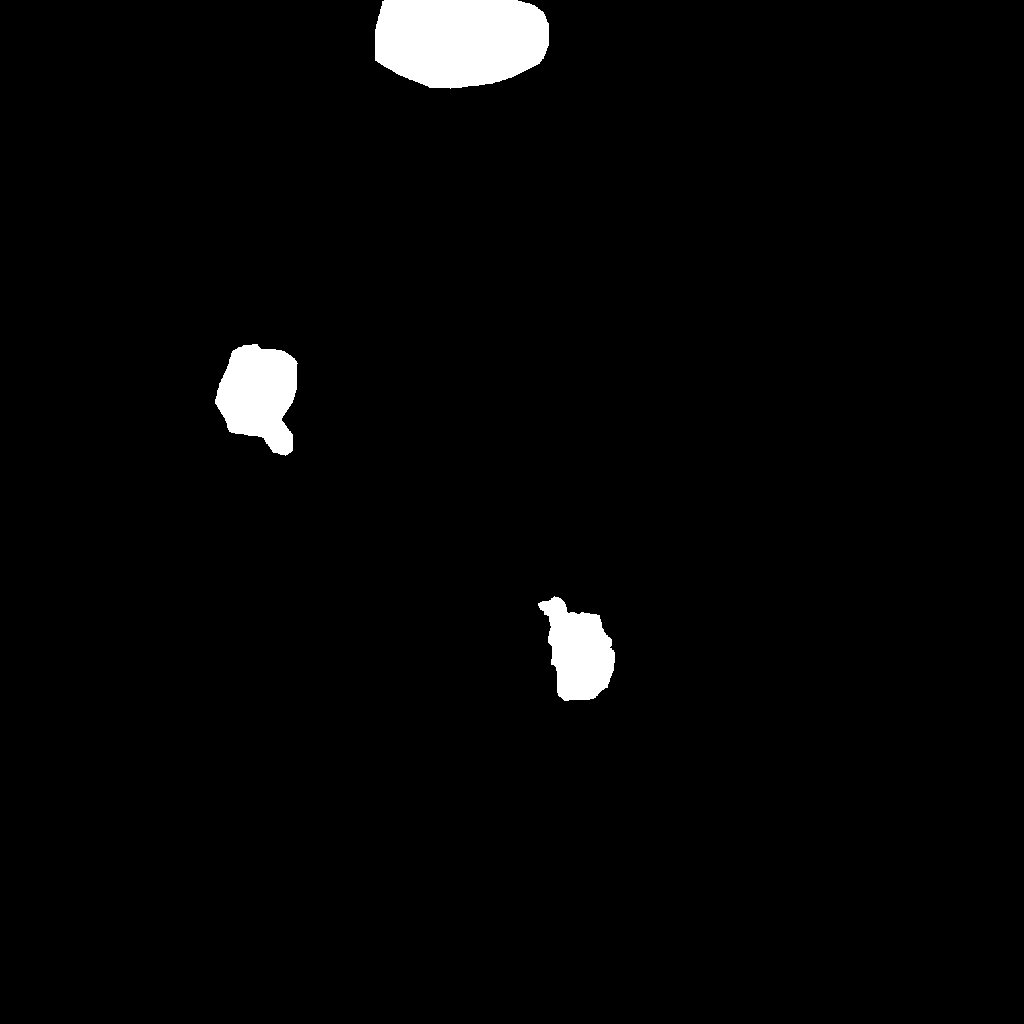} &
\includegraphics[width=\mywidth,  ,valign=m, keepaspectratio,] {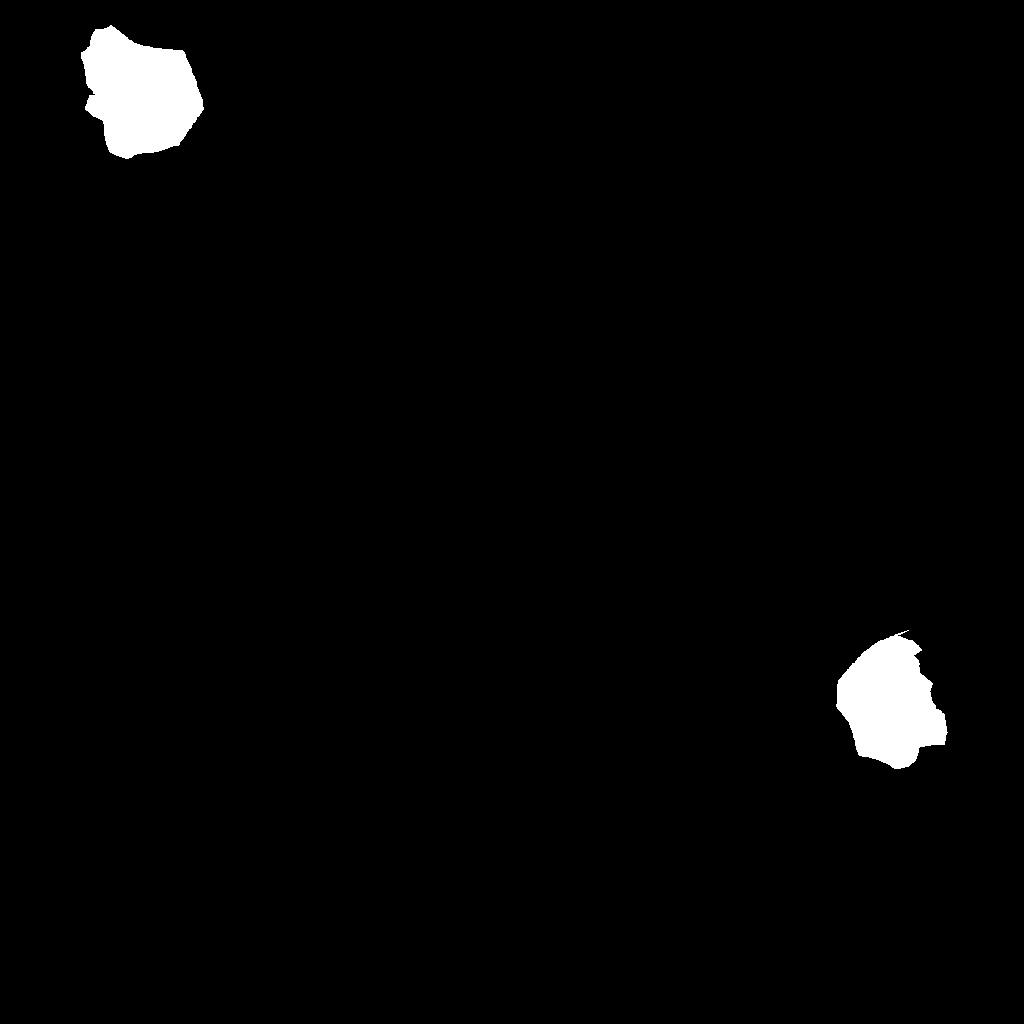} 

\\

& \tiny{capsules} & \tiny{cashew} & \tiny{chewinggum} & \tiny{fryum} & \tiny{macaroni1} & \tiny{macaroni2} & \tiny{pcb1}& \tiny{pcb3} & \tiny{pcb4} & \tiny{pipe fryum} \\

{\rotatebox[origin=t]{90}{\textit{\textbf{\tiny I}}}} & 
\includegraphics[width=\mywidth,  ,valign=m, keepaspectratio,] {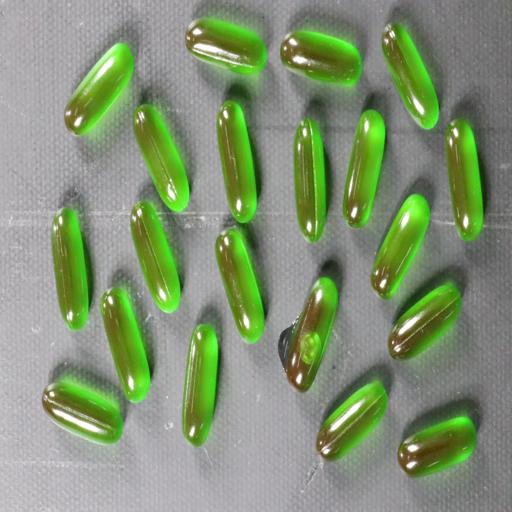} &
\includegraphics[width=\mywidth,  ,valign=m, keepaspectratio,] {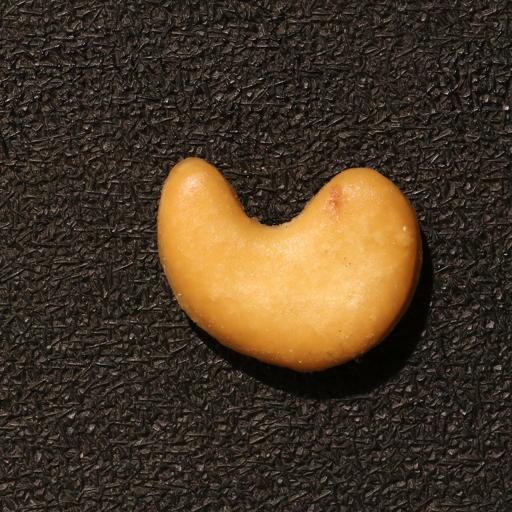} &
\includegraphics[width=\mywidth,  ,valign=m, keepaspectratio,] {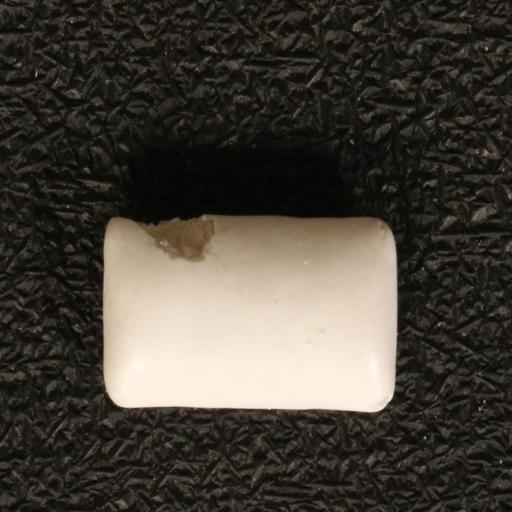} &
\includegraphics[width=\mywidth,  ,valign=m, keepaspectratio,] {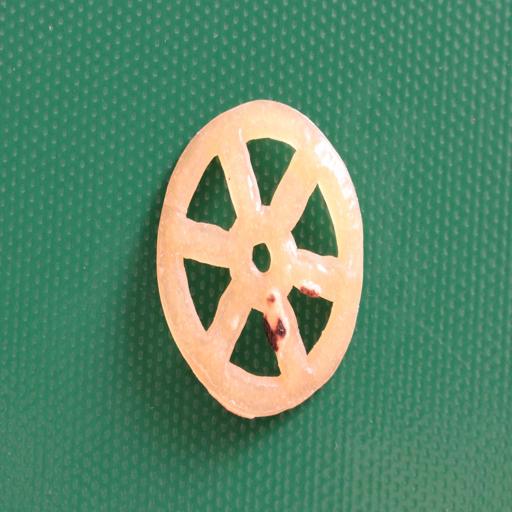} &
\includegraphics[width=\mywidth,  ,valign=m, keepaspectratio,] {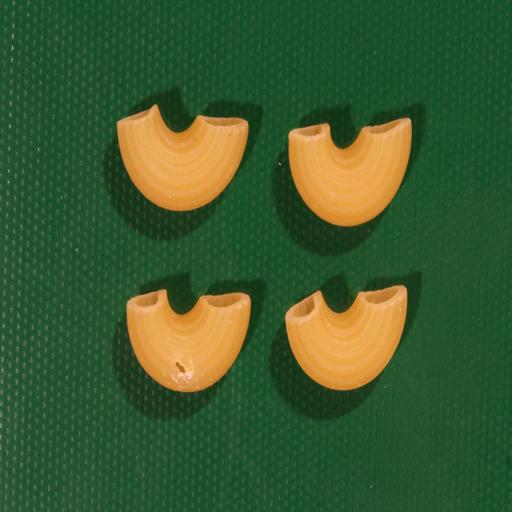} &
\includegraphics[width=\mywidth,  ,valign=m, keepaspectratio,] {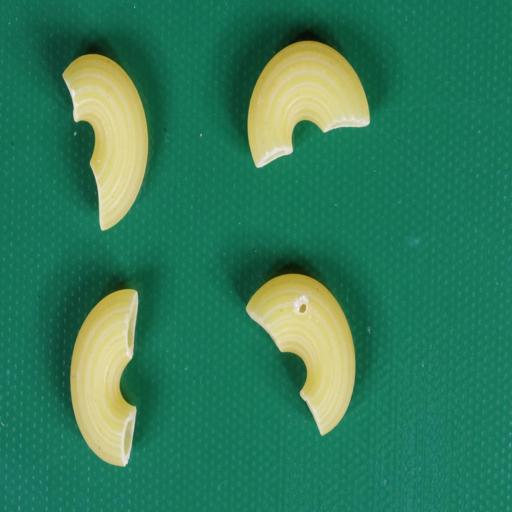} &
\includegraphics[width=\mywidth,  ,valign=m, keepaspectratio,] {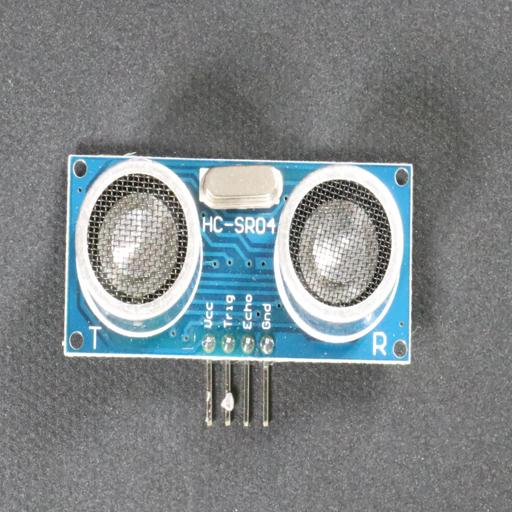} &
\includegraphics[width=\mywidth,  ,valign=m, keepaspectratio,] {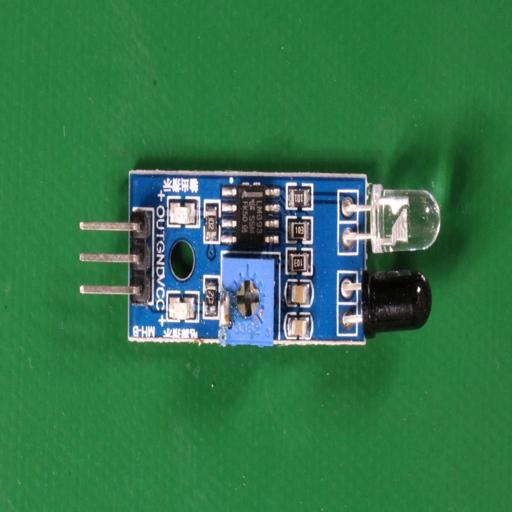} &
\includegraphics[width=\mywidth,  ,valign=m, keepaspectratio,] {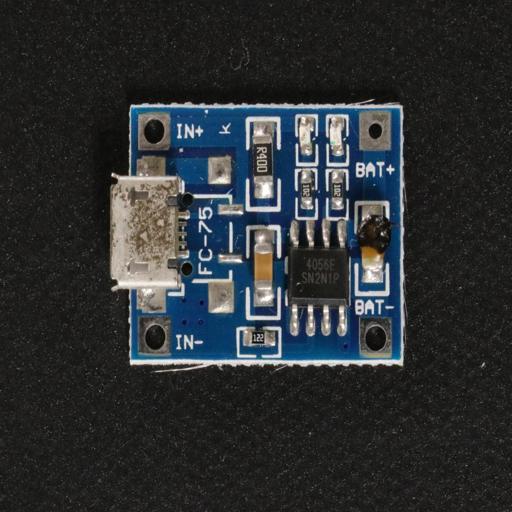} &
\includegraphics[width=\mywidth,  ,valign=m, keepaspectratio,] {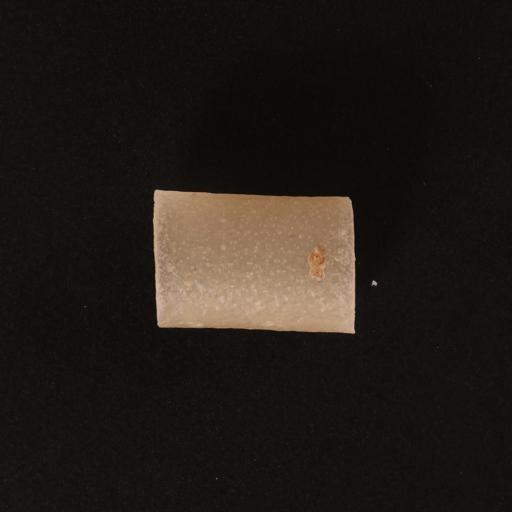} 
\\

{\rotatebox[origin=t]{90}{\textit{\textbf{ \tiny Inv.}}}}&  
\includegraphics[width=\mywidth,  ,valign=m, keepaspectratio,] {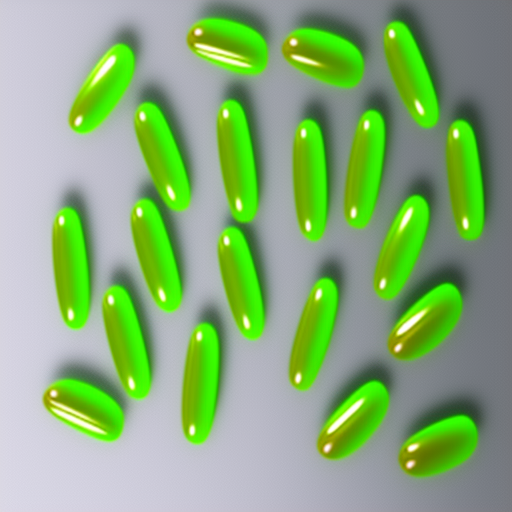} &
\includegraphics[width=\mywidth,  ,valign=m, keepaspectratio,] {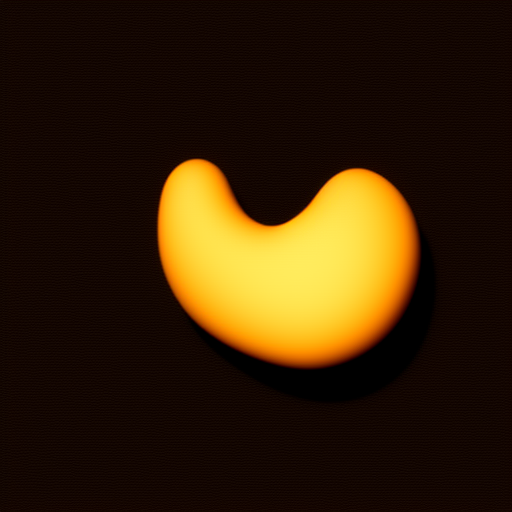} &
\includegraphics[width=\mywidth,  ,valign=m, keepaspectratio,] {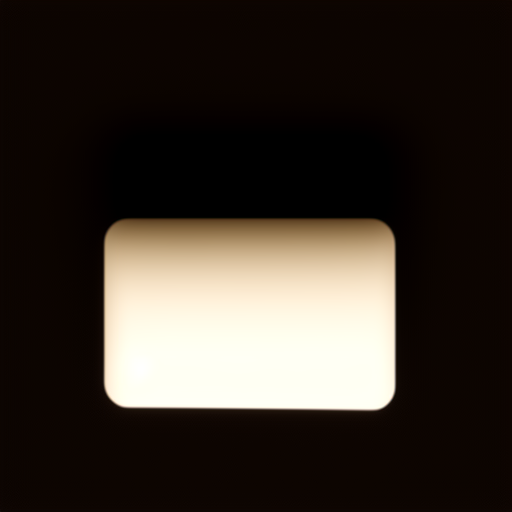} &
\includegraphics[width=\mywidth,  ,valign=m, keepaspectratio,] {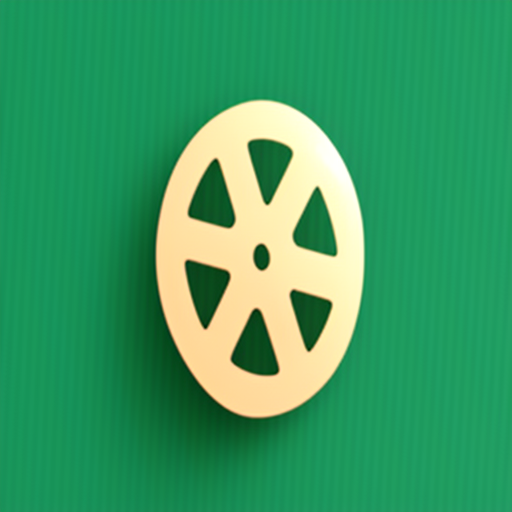} &
\includegraphics[width=\mywidth,  ,valign=m, keepaspectratio,] {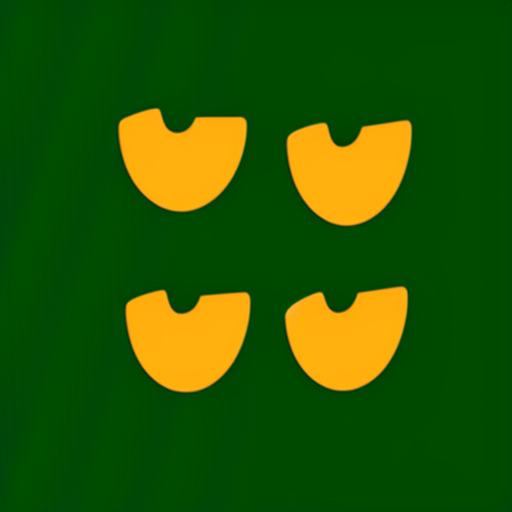} &
\includegraphics[width=\mywidth,  ,valign=m, keepaspectratio,] {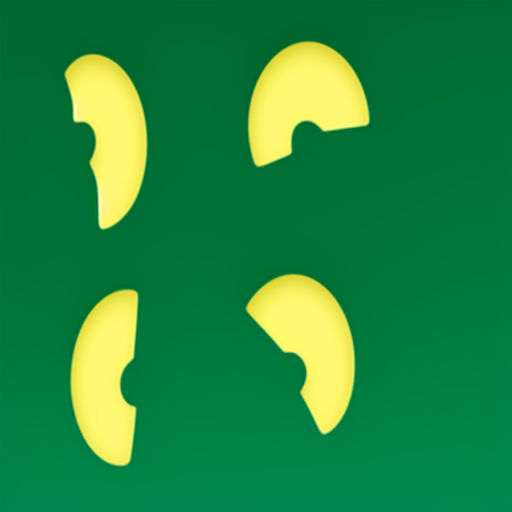} &
\includegraphics[width=\mywidth,  ,valign=m, keepaspectratio,] {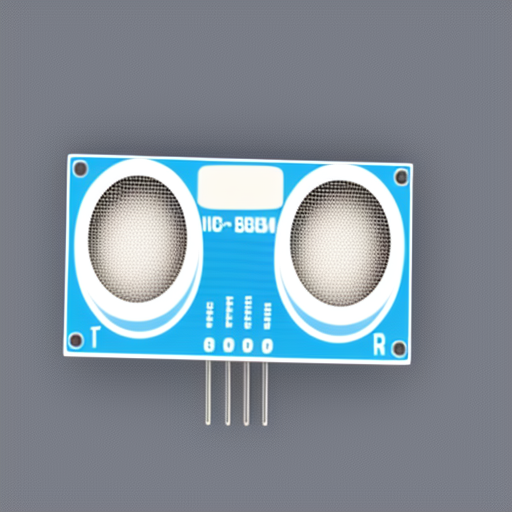} &
\includegraphics[width=\mywidth,  ,valign=m, keepaspectratio,] {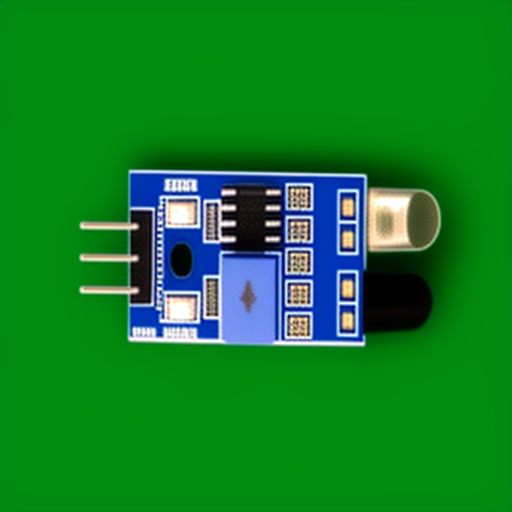} &
\includegraphics[width=\mywidth,  ,valign=m, keepaspectratio,] {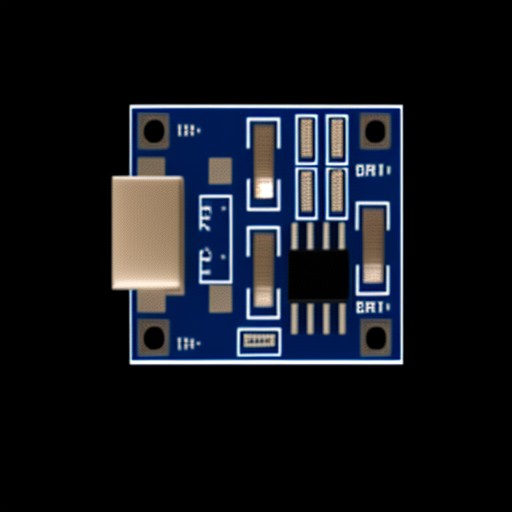} &
\includegraphics[width=\mywidth,  ,valign=m, keepaspectratio,] {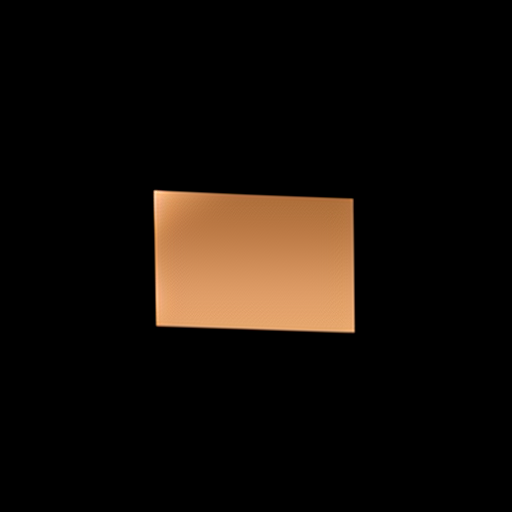} 
\\

{\rotatebox[origin=t]{90}{\textit{\textbf{\tiny {Pred.}}}}} & 
\includegraphics[width=\mywidth,  ,valign=m, keepaspectratio,] {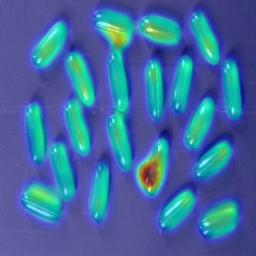} &
\includegraphics[width=\mywidth,  ,valign=m, keepaspectratio,] {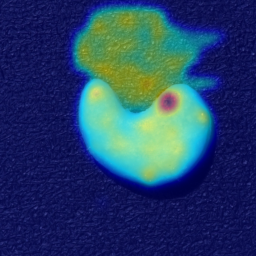} &
\includegraphics[width=\mywidth,  ,valign=m, keepaspectratio,] {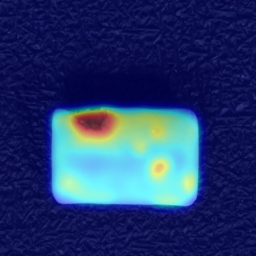} &
\includegraphics[width=\mywidth,  ,valign=m, keepaspectratio,] {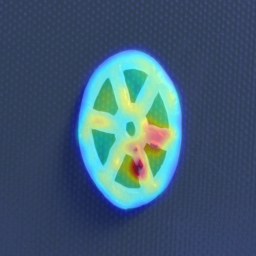} &
\includegraphics[width=\mywidth,  ,valign=m, keepaspectratio,] {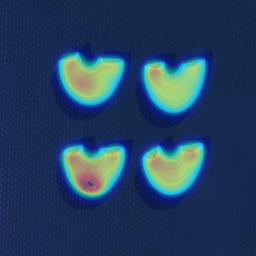} &
\includegraphics[width=\mywidth,  ,valign=m, keepaspectratio,] {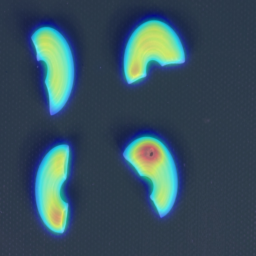} &
\includegraphics[width=\mywidth,  ,valign=m, keepaspectratio,] {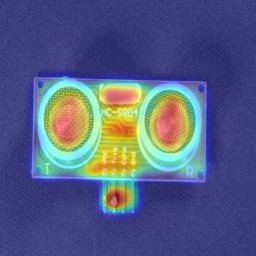} &
\includegraphics[width=\mywidth,  ,valign=m, keepaspectratio,] {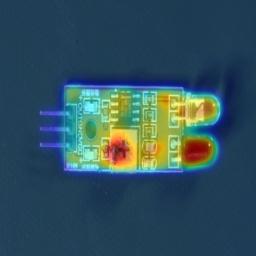} &
\includegraphics[width=\mywidth,  ,valign=m, keepaspectratio,] {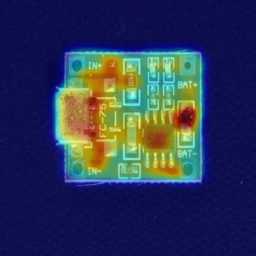} &
\includegraphics[width=\mywidth,  ,valign=m, keepaspectratio,] {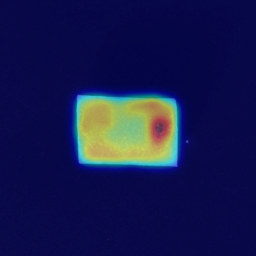}  
\\

{\rotatebox[origin=t]{90}{\textit{\textbf{\tiny GT}}}} & 
\includegraphics[width=\mywidth,  ,valign=m, keepaspectratio,] {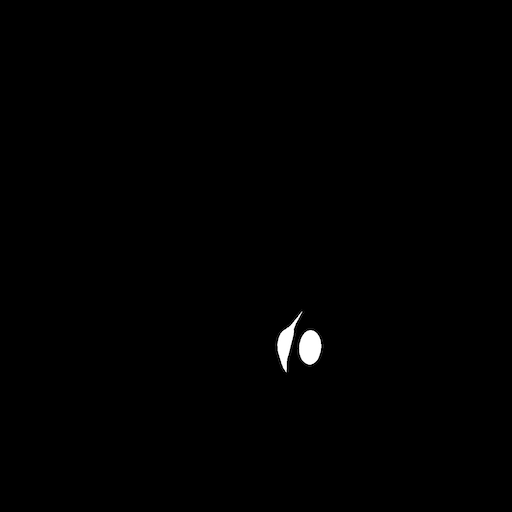} &
\includegraphics[width=\mywidth,  ,valign=m, keepaspectratio,] {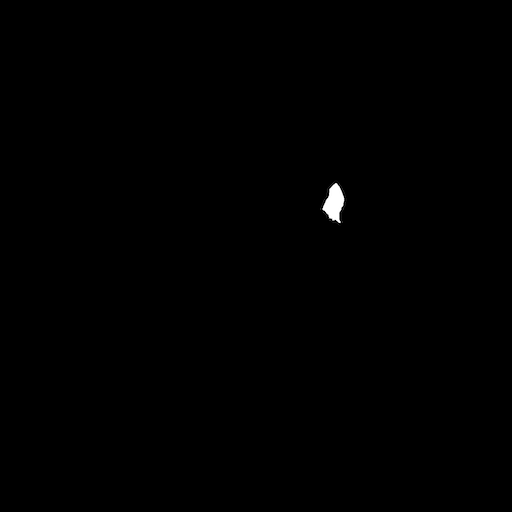} &
\includegraphics[width=\mywidth,  ,valign=m, keepaspectratio,] {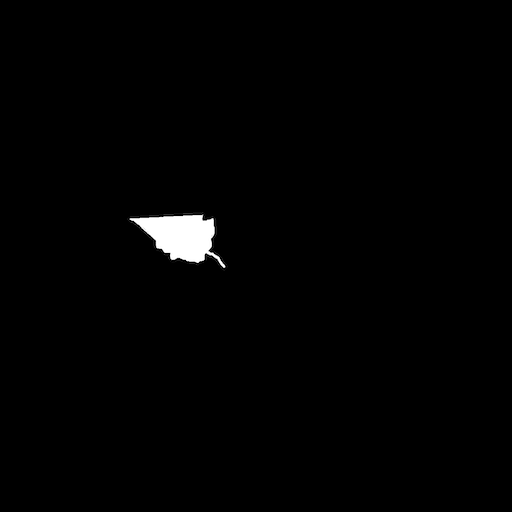} &
\includegraphics[width=\mywidth,  ,valign=m, keepaspectratio,] {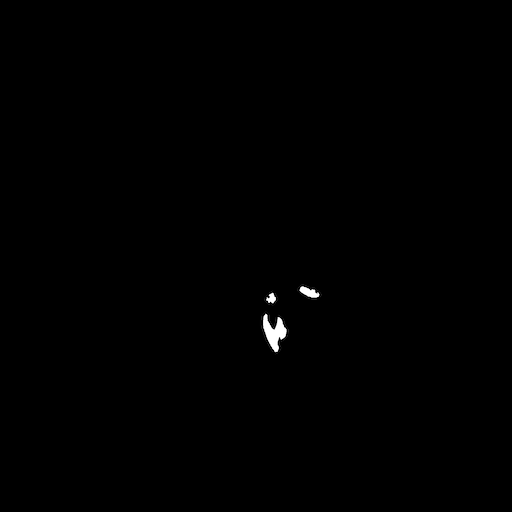} &
\includegraphics[width=\mywidth,  ,valign=m, keepaspectratio,] {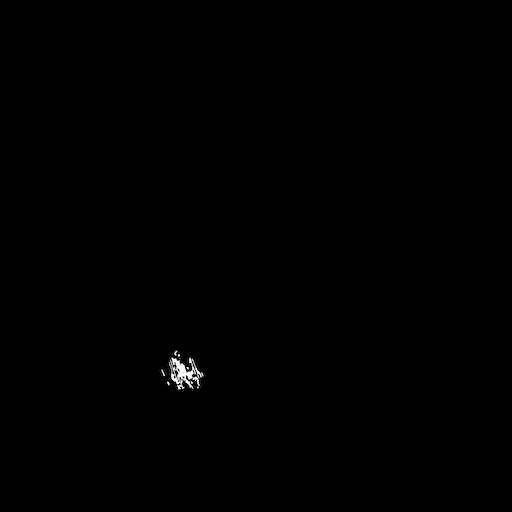} &
\includegraphics[width=\mywidth,  ,valign=m, keepaspectratio,] {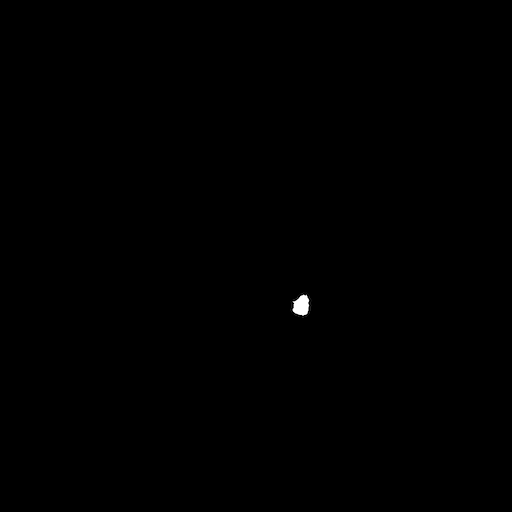} &
\includegraphics[width=\mywidth,  ,valign=m, keepaspectratio,] {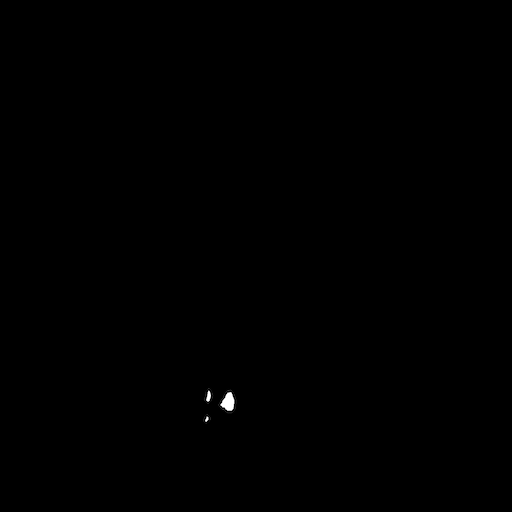} &
\includegraphics[width=\mywidth,  ,valign=m, keepaspectratio,] {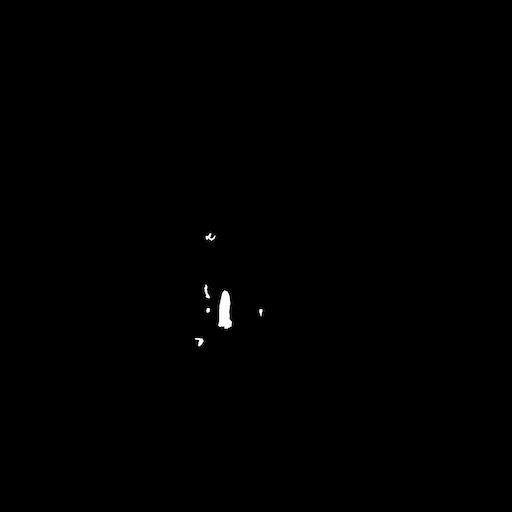} &
\includegraphics[width=\mywidth,  ,valign=m, keepaspectratio,] {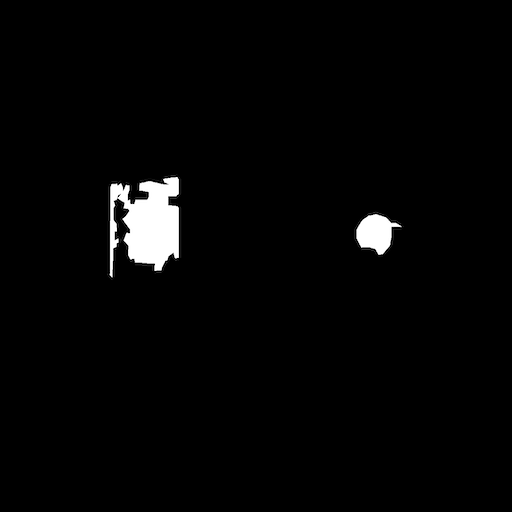} &
\includegraphics[width=\mywidth,  ,valign=m, keepaspectratio,] {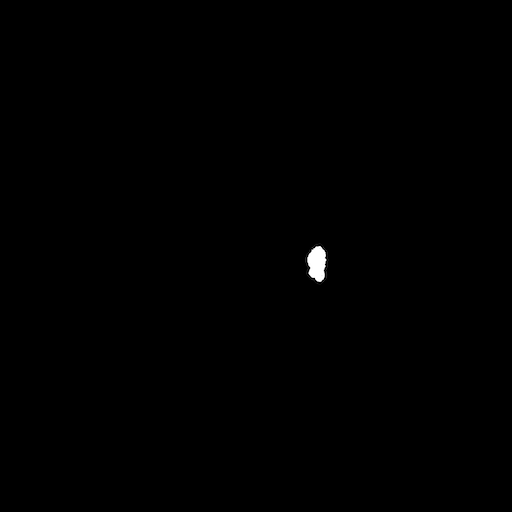}

\end{tabular}}
\caption{Qualitative results of {\ours} on MVTec-AD (top) and VISA (bottom) for selected classes. {\ours} successfully detects the anomaly locations on various object and texture classes.}
\label{fig:visual_results_mvtec}
% \vspace{-5mm}
\end{figure*}

\subsection{\ours}

We present the detailed structure of our method in Figure~\ref{fig:model}. As our method is training-free, we use only the test images of each dataset. The inference pipeline starts by feeding the input test image $I$ to the SD image encoder $\mathcal{E}_I$ of the pretrained SD model. The image encoder extracts the latent code $z$ for the input image. Then, the backward diffusion is applied $T$ times in reverse to add noise to $z$ to obtain $z_T$ with a simple text description of the image in the form of \textit{"an image of a [obj]"}. Here \textit{obj} corresponds to the class label of the input image. This noisy latent corresponds to a vector which is responsible for generating the input image $I$. When the denoising process is applied $T$ times to the $z_T$, normally we should end up with $z$. However, as the network introduces error during the reverse process and the employed classifier guidance, usually denoised $z_T$ and $z$ do not exactly match. Hence, there is a difference between the input image $I$ and the generated image $I_T$ from $z_T$. A common approach to achieve almost the same reconstruction is to start the inversion from a small $T'$. The noisy latent $z_{T'}$ is fed to the SD model for denoising again for $T$ steps to obtain $\tilde{z}$. The denoised latent $\tilde{z}$ is decoded back to the image space using the pretrained SD image decoder $\mathcal{D}_I$ to get the reconstructed image $\tilde{I}$. In order to segment the anomaly locations, the input image $I$ and the reconstructed image $\tilde{I}$ are fed to an off-the-shelf self-supervised model (i.e., DINO~\cite{dino}) to extract the features $F$ and $\tilde{F}$, respectively. Finally, the pixels with high dissimilarity between these two feature vectors denote the Anomaly Map.

\begin{table}
\caption{Details of the MVTec-AD and VISA test sets. Obj. stands for Object and Tex. stands for Texture categories.}

\centering
\setlength\tabcolsep{4.0pt}
\begin{tabular}{cccccc}
\toprule
\multirow{2}{*}{Dataset} & \multicolumn{2}{c}{Categories} & & \multicolumn{2}{c}{Images}  \\
\cmidrule(lr){2-3} \cmidrule(lr){5-6} 
& {Obj.} & {Tex.} &  & Anomaly & Normal\\
\toprule
MVTec-AD~\cite{bergmann2019mvtec}   & 10 &  5 & & 1,258 & 467\\ 
VisA~\cite{visa}                & 12 &	\na & & 962   & 1,200  \\
\bottomrule
\end{tabular}%}
\label{tab:dataset_mvtec_visa}
% \vspace{-3mm}
\end{table}

We observe that for some classes, there are irregularities in the background itself which need to be filtered out. To this end, we utilise a pretrained unsupervised image segmentation model CutLER~\cite{wang2023cut} to extract the binary object mask(s) $M_{Obj}$ of the input image $I$. The estimated mask is multiplied by the predicted anomaly map $A$ to get the final segmentation mask.

\section{Experiments}
\label{sec:exp}

%%%% Table MVTEC-VISA Results.
\begin{table*}[t]
\caption{Comparison of \textbf{ZSAD methods} on the \textbf{MVTec-AD}~\cite{bergmann2019mvtec} and \textbf{VISA}~\cite{visa} datasets.~$\dagger$ denotes methods without publicly available implementation. {\na} means there is no evaluation on the specific dataset. In the \textbf{training-free} category, our method achieves on-par results compared to guided-prompt based methods on MVTec-AD, and outperforms all training-free ZSAD methods on pixel-level metrics on VISA dataset.}
\label{tab:mvtec_visa_scores}

\small
\begin{center}
\resizebox{1.0\textwidth}{!}{
\begin{tabular}{lccccccc|ccccc}
\toprule % <-- Toprule here
\multirow{3}{*}{Method} & \multirow{3}{*}{Train} & \multirow{3}{*}{\shortstack{Guided \\Prompts}}&  \multicolumn{5}{c}{MVTec-AD} & \multicolumn{5}{c}{VISA}\\
\cmidrule(lr){4-8} \cmidrule(lr){9-13}
 & & &  $ROC_{I}$ & $ROC_{P}$ &  {PRO} & $AP_P$ & $F1_P$ &  $ROC_{I}$ & $ROC_{P}$ &  {PRO} & $AP_P$ & $F1_P$ \\
\midrule % <-- Midrule here
APRIL-GAN~\cite{chen2023april} & \Checkmark & \Checkmark& 86.1  & 87.6 & 44.0 & 40.8 & 43.3 & 77.5  & 94.2 & 86.6 & 25.8 & 32.3  \\
AnomalyCLIP~\cite{zhouanomalyclip} & \Checkmark & \Checkmark& 91.5  & 91.1 & 81.4 & 34.5 & 39.1 & 80.9  & 95.5 & 86.7 & 21.3 & 28.3 \\
FiLo~\cite{gu2024filo} & \Checkmark & \Checkmark& 91.4  & \textbf{93.1} & 61.7 & \textbf{52.2} & \textbf{52.0 } & 83.9  & \textbf{95.9} & 85.4 & 31.4 & \textbf{38.5}\\
PromptAD~\cite{li2024promptad}~$\dagger$  & \Checkmark & \Checkmark& 90.8  & 92.1 & 72.8 & - & 36.2  & \na  & \na & \na & \na & \na  \\
AdaCLIP~\cite{cao2025adaclip} & \Checkmark & \Checkmark& 89.9  & 89.9 & 44.0 & 41.6 & 43.9 & 86.2  & 95.8 & 51.0 & \textbf{31.5} & 37.1 \\
Bayes-PFL~\cite{Qu_2025_CVPR} & \Checkmark & \Checkmark& 92.3  & 91.8 & \textbf{87.4} & 48.3 & - & 87.0  & 95.6 & \textbf{88.9} & 29.8 & -  \\
DZAD~\cite{zhang2025dzad}~$\dagger$ & \Checkmark & \XSolidBrush & \textbf{93.5}  & 86.7 & - & - & - & \textbf{90.2}  & 92.0 & - & - & - \\
\midrule % <-- Midrule here
WinCLIP~\cite{winclip} & \XSolidBrush & \Checkmark& 91.8  & 85.1 & 64.6 & 18.2 & 31.6 & 78.1  & 79.6 & 56.8 & 5.4 & 14.8  \\
ALFA~\cite{zhu2024llms}~$\dagger$ & \XSolidBrush & \Checkmark& \textbf{93.2}  & \textbf{90.6} & 78.9 & - & 36.6 & 81.2  & 85.9 & 63.8 & - & 15.9 \\
FADE~\cite{lifade} & \XSolidBrush & \Checkmark& 90.0  & 89.6 & \textbf{84.5} & \textbf{34.1} & \textbf{39.8} & 78.1  & 91.5 & \textbf{79.3} & 10.0 & 16.7 \\
AnomalyVLM~\cite{cao2025personalizing} & \XSolidBrush & \Checkmark& 62.4  & 73.1 & 42.4 & 28.7 & 37.7  & 71.0  & 75.8 & 36.9 & 13.8 & 22.1 \\
RAG~\cite{rag_zsad}~$\dagger$ & \XSolidBrush & \Checkmark& 92.5 & 89.1 & - & - & - & \textbf{86.8} & 87.6 & - & - & - \\
MAEDAY~\cite{maeday}~$\dagger$ & \XSolidBrush & \XSolidBrush & 74.5  & 69.4 & - & - & - & \na  & \na & \na & \na & \na \\
\textbf{{\ours} (ours)} & \XSolidBrush & \XSolidBrush& 76.0&  88.6& 73.3 & 31.6 & 35.7 & 69.7 & \textbf{93.4} & 78.2 & \textbf{19.5} & \textbf{24.0}\\
% % \textbf{{\ours} (ours)} + 1 Anchor & \XSolidBrush & \XSolidBrush& 80.7 &  91.3 & 78.1 & 41.4 & 44.1\\
\bottomrule % <-- Bottomrule here
\end{tabular}}
\end{center}
% \vspace{-5mm}
\end{table*}

\subsection{Datasets}

MVTec-AD contains $10$ "object" and $5$ "texture" classes, with 1258 anomaly and 467 normal test images. The VISA dataset contains only "object" classes and has 962 anomaly and 1200 normal test images. Note that the VISA test split has a balanced number of normal and anomaly images. Since our method is training-free, we are not interested in the training statistics of the datasets. Details of both datasets are presented in Table~\ref{tab:dataset_mvtec_visa}.

\subsection{Experimental Setup}
\label{sec:setup}

We use the publicly available Stable Diffusion $2.1$ weights~\footnote{huggingface.co/stabilityai/stable-diffusion-2-1} for our pipeline. We use $50$ for the reverse sampling timestep $T$, and $10$ as the $T'$ for the diffusion denoising starting point. For all datasets, images are resized to $256\times256$ pixels after reconstruction for inference. All the experiments are performed on a single Nvidia A100 GPU. 

We use DINO ViTS/8~\cite{dino} as the feature extractor. We first extract patch features from the last layer for the input and reconstructed images, and then measure the cosine dissimilarity (i.e. $1- cossim( \tilde{F}, F)$) between the patch features. The final heatmap is resized to the input size to locate the anomalies.

For object segmentation, we use an unsupervised model CutLER~\cite{wang2023cut} with Cascade Mask R-CNN as the backbone detector. We set the confidence threshold as $0.1$ and merge all the pixels above this threshold to use as object masks.

We run our method for $5$ times on all datasets and observed only fractional differences which could be accounted for floating point operations. 
Our method consumes less than $7.5$ GB of GPU memory and takes $5$ seconds to find the inverted latents and another $5$ seconds to denoise the noisy latents on a single NVIDIA RTX A6000 GPU.

\subsection{Qualitative Results}
\label{sec:visual_res}

We show visual results of {\ours} in Figure~\ref{fig:visual_results_mvtec} for the MVTec-AD (top) and VISA (bottom) datasets, respectively.

On MVTec-AD, our method effectively segments both object and texture categories across diverse defect types, including \textit{broken}, \textit{dent}, \textit{color}, and \textit{bent}. As illustrated in Figure~\ref{fig:visual_results_mvtec} (top), representative examples include the \textit{bottle}, \textit{metal nut}, \textit{carpet}, and \textit{transistor} categories.
Even large (e.g., \textit{bottle, cable}) and multiple anomalies (e.g., \textit{grid, wood, zipper}) are localised with high precision. 

On VISA dataset, anomalies are often more subtle and appear in fine-grained details, making the detection task more challenging. Nonetheless, {\ours} demonstrates strong performance across various object categories with minor anomalies, such as \textit{cashew}, \textit{fryum}, and \textit{macaroni}.

\subsection{Quantitative Results}
\label{sec:numeric_res}

\textbf{Metrics.} Following common practice, we used five metrics: Mean Area under the ROC curve mAU-ROC~\cite{zavrtanik2021draem} on image and pixel levels, pixel-level per-region-overlap $PRO$~\cite{Bergmann_Fauser_Sattlegger_Steger_2020} score, $F1_P$, and average precision $AP_P$.

% \begin{table}
\begin{table}
\caption{Comparison of \textbf{training-free ZSAD methods} on the \textbf{MPDD}~\cite{mpdd} dataset. Our method achieves SoTA results.}
\label{tab:mpdd_btad_results}
\begin{center}
\resizebox{1.0\columnwidth}{!}{
\begin{tabular}{lccccc}
\toprule % <-- Toprule here
% \multirow{3}{*}{Method}   & \multicolumn{5}{c}{MPDD}  \\ 
%   \cmidrule(lr){2-6}
& $ROC_{I}$ & $ROC_{P}$ &  $PRO$ & $AP_{P}$ & $F1_{P}$ \\
\midrule % <-- Midrule here
% AdaClip & 1  & 2 & 3 & 1 & 2 & 3  \\
% \midrule % <-- Midrule here
% WinCLIP & 61.4   & 71.2 & 15.4 & 68.2 & 72.6 & 18.5  \\
% FADE & 62.8   & 94.2 & 24.6 & 74.3 & \textbf{96.3 }& \textbf{46.4}  \\
% AnomalyVLM & 42.7  & 81.7 & 18.9 & 59.0 & 65.8 & 14.8  \\
% \textbf{{\ours}} & \textbf{63.4}  & \textbf{94.9} & \textbf{27.0} & \textbf{79.8} & 68.4 & 19.1  \\

WinCLIP~\cite{winclip} & 61.4  & 71.2 & 53.4 & 6.7 & 15.4   \\
FADE~\cite{lifade} & 62.8  & 94.2 & 81.9 & 18.9 & 24.6  \\
AnomalyVLM~\cite{cao2025personalizing} & 50.5  & 78.0 & 27.8 & 7.5 & 18.1 \\
\textbf{{\ours}} & \textbf{63.4}  & \textbf{94.9} & \textbf{82.0} & \textbf{22.9} & \textbf{27.0}  \\
\bottomrule % <-- Bottomrule here
\end{tabular}}
\end{center}
\end{table}

\noindent\textbf{State-of-the-art comparison.} {\ours} is compared with both training-based and training-free ZSAD methods. If the value for a metric is present in the original paper or official repository, we directly use it. When available, we use official code or pretrained models to calculate the missing values.

%%%%%%%%%%%% Example Failure cases
\begin{figure}
\captionsetup[subfigure]{labelformat=empty}
\centering
\setlength\tabcolsep{1.5pt} % default value: 6pt
\resizebox{1.0\linewidth}{!}{
\begin{tabular}{c@{}c@{}c@{}c}
{\textit{\textbf{\tiny{Image}}}} & {\textit{\textbf{\tiny{Inv}}}} &  {\textit{\textbf{\tiny{Pred}}}} & {\textit{\textbf{\tiny{GT}}}}  \\
\includegraphics[width=\myw,  ,valign=m, keepaspectratio,] {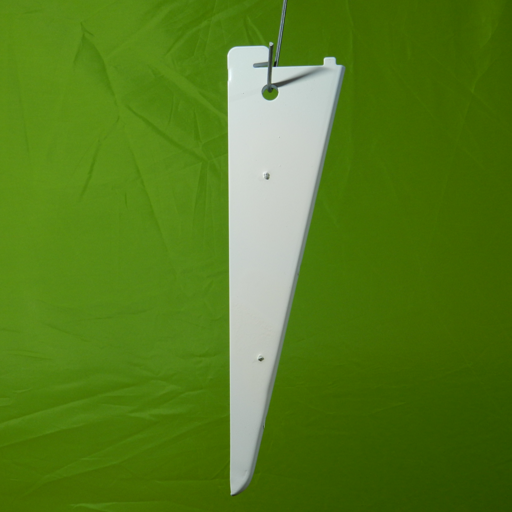} &
\includegraphics[width=\myw,  ,valign=m, keepaspectratio,] {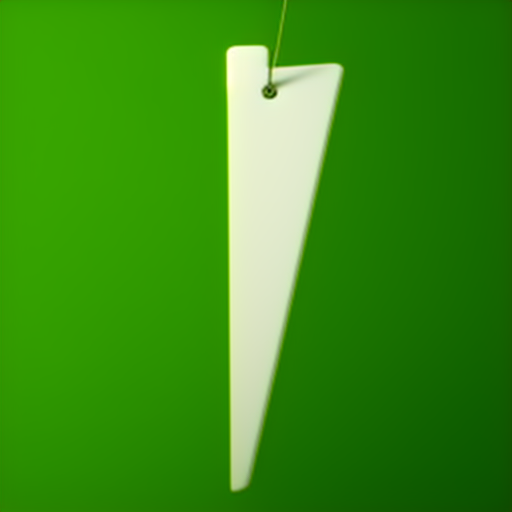}  &
\includegraphics[width=\myw,  ,valign=m, keepaspectratio,] {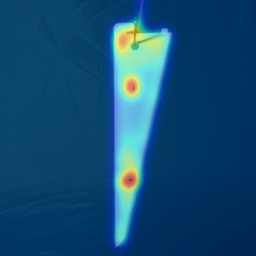}  &
\includegraphics[width=\myw,  ,valign=m, keepaspectratio,] {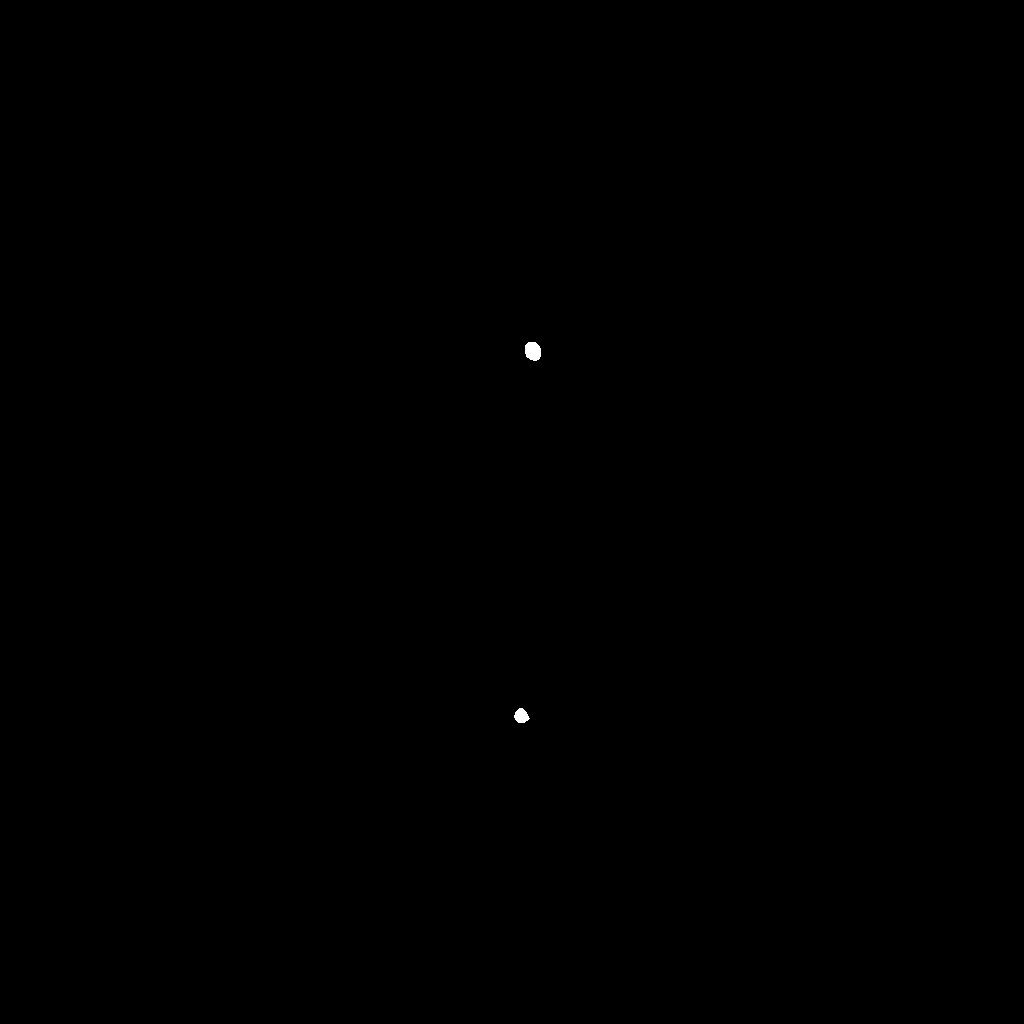}  \\

\includegraphics[width=\myw,  ,valign=m, keepaspectratio,] {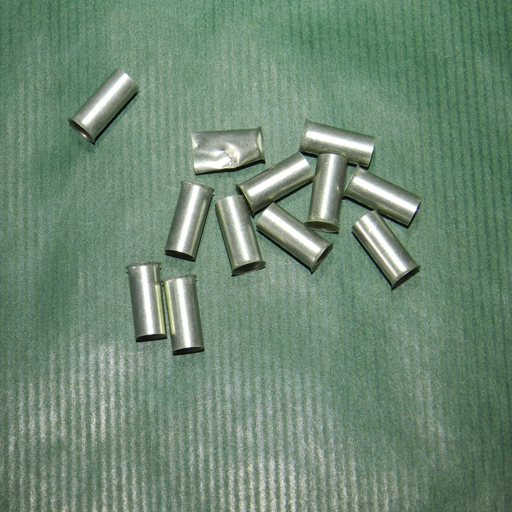} &
\includegraphics[width=\myw,  ,valign=m, keepaspectratio,] {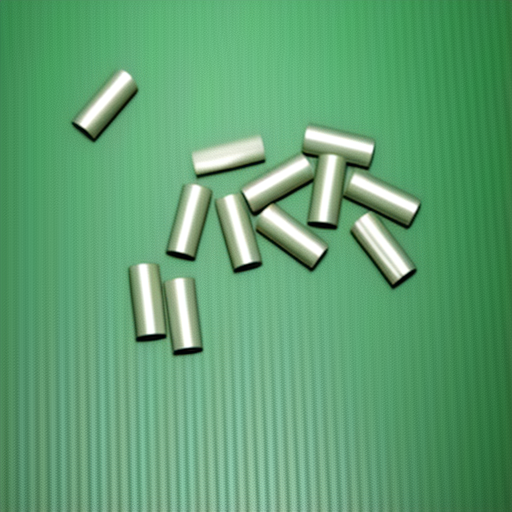}  &
\includegraphics[width=\myw,  ,valign=m, keepaspectratio,] {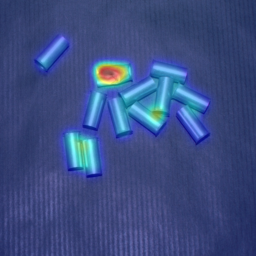} &
\includegraphics[width=\myw,  ,valign=m, keepaspectratio,] {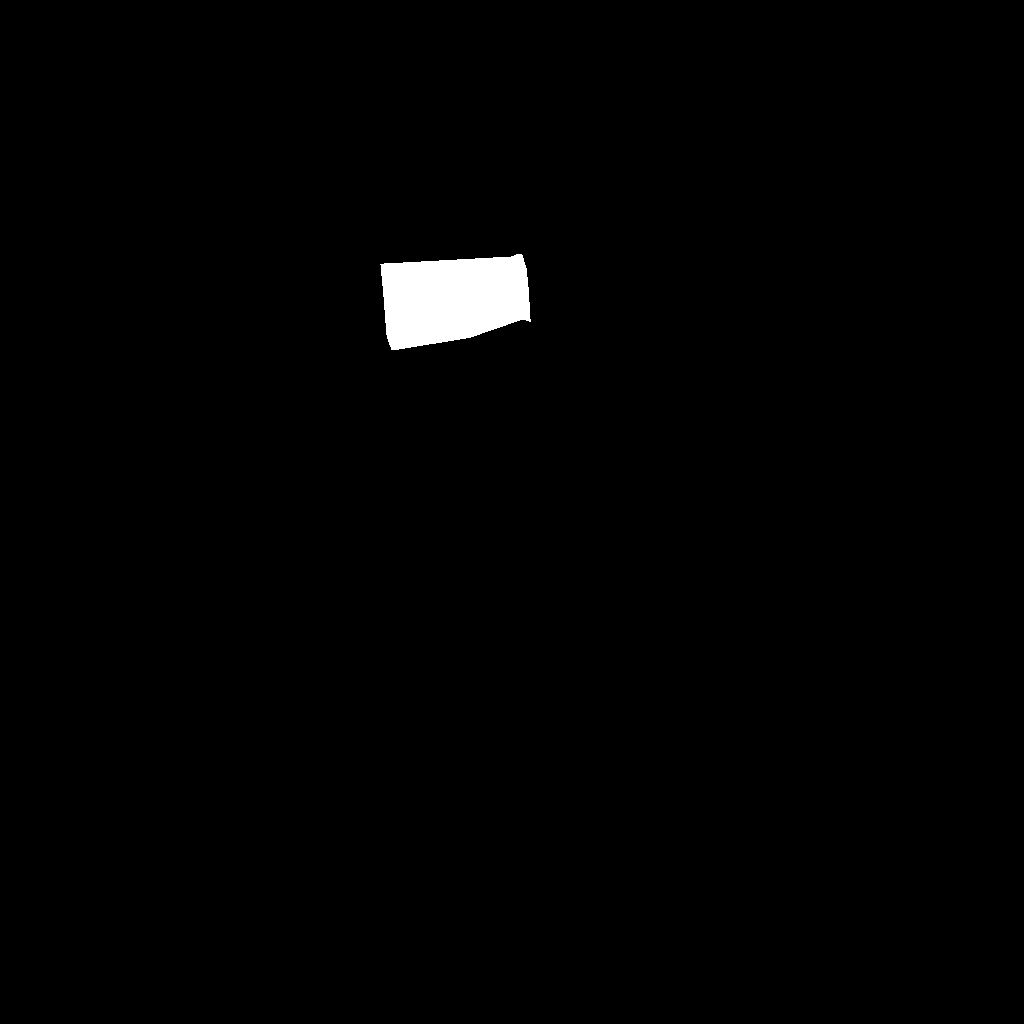} \\
\end{tabular}}
\caption{Examples of \ours{} predictions on the MPDD dataset.}
\label{fig:mpdd_res_main}
\end{figure}

\noindent\textbf{MVTec-AD and VISA} quantitative results are presented in Table~\ref{tab:mvtec_visa_scores}. {{\ours}} achieves on-par results compared to training-free ZSAD methods on the MVTec-AD dataset. Similarly, on the VISA dataset, our method establishes state-of-the-art performance for pixel-level $ROC$, $AP$, and $F_1$ among training-free ZSAD methods. Furthermore, compared to the previous diffusion-based method DZAD~\cite{zhang2025dzad} (the best performing on the image-level ROC among the training-based methods), {\ours} achieves better pixel-level scores without any training.

\noindent\textbf{MPDD.} In order to demonstrate the generalisation of {\ours}, we experimented with another challenging industrial defect dataset. We compare our method only to training-free ZSAD methods (c.f. Table~\ref{tab:mpdd_btad_results}), and {\ours} significantly improves the current state-of-the-art, especially on the pixel-level metrics. We present visual results of {\ours} in Figure~\ref{fig:mpdd_res_main} for selected classes. 
Detailed class-wise results on all datasets are available in the Supplementary Material.

\subsection{Ablation Experiments}
\label{sec:abl_exp}

\noindent\textbf{Effect of using object mask extraction (c.f. Table~\ref{tab:abl_on_mask}).} 
In order to see the effect of using the extracted object masks ($M_{Obj}$), we calculate the performance of {\ours} with and without this step. 
As MVTec-AD contains both texture and object classes, we run separate experiments by enabling the mask extraction for each group.

\noindent On the VISA dataset, where there are only objects, using the object mask to refine the anomaly maps brings dramatic improvements. This is due to the fact that the background on this dataset has some irregularities, and they are detected as false positives by {\ours}. 

\begin{table}[t]
\caption{Effect of using object segmentation method for object and texture classes.}
\label{tab:abl_on_mask}

\begin{center}
\resizebox{1.0\columnwidth}{!}{
\begin{tabular}{ccc|ccccc}
 \toprule % <-- Toprule here
\multirow{3}{*}{Dataset} &  \multicolumn{2}{c}{Obj. Mask}   &  \multicolumn{5}{c}{Metrics} \\
 \cmidrule(lr){2-3}  \cmidrule(lr){4-8}
  & Obj. &  Tex.  & $ROC_I$ & $ROC_P$ & $PRO$ & $AP_P$ & $F1_P$ \\
\midrule % <-- Midrule here
\multirow{3}{*}{MVTec-AD} &  \XSolidBrush & \XSolidBrush  & \textbf{77.5}      &  85.7 & \textbf{76.8} & \textbf{33.5} & \textbf{36.5}      \\
 &  \Checkmark & \XSolidBrush & 76.0     &  \textbf{88.6} & 73.3 & 31.6 & 35.7      \\
 &  \Checkmark & \Checkmark  & 70.9      &  79.9 & 57.6 & 23.0 & 29.1       \\
\midrule
\multirow{2}{*}{VISA} &   \XSolidBrush & \na   & 67.5 & 85.7 & 67.6 & 13.0 & 17.9    \\
&   \Checkmark & \na   & \textbf{69.7} & \textbf{93.4} & \textbf{78.2} & \textbf{19.5} & \textbf{24.0}    \\
\bottomrule % <-- Bottomrule here
\end{tabular}}
\end{center}
% \vspace{-7mm}
\end{table}

\begin{table*}
\caption{Ablation experiments on the effect of varying the timestep and SD version.}
\label{tab:abl_on_timestep_sd}
\begin{center}
% \resizebox{1.0\linewidth}{!}{
\begin{tabular}{cc|ccccc|ccccc}
 \toprule % <-- Toprule here
\multirow{3}{*}{\shortstack{Time\\step}} &  \multirow{3}{*}{\shortstack{SD\\vers.}}   &  \multicolumn{5}{c}{MVTec-AD} &  \multicolumn{5}{c}{VISA} \\
 \cmidrule(lr){3-7} \cmidrule(lr){8-12} 
 &  & $ROC_I$ & $ROC_P$ & $PRO$ & $AP_P$ & $F1_P$ & $ROC_I$ & $ROC_P$ & $PRO$ & $AP_P$ & $F1_P$  \\
\midrule % <-- Midrule here
30 & 1.5  &  70.9  & 87.7 & 71.1 & 28.0 & 33.6 & 64.0 & 92.4 & 74.3 & 15.6 & 19.2   \\
20 & 1.5  &  72.2  & 87.9 & 72.5 & 30.0 & 34.8 & 69.1 & 93.4 & 77.8 & 18.4 & 23.4   \\
15 & 1.5  &  74.2  & 87.9 & 72.5 & 30.7 & 35.2 & 69.4 & 93.4 & 78.1 & 19.5 & 24.1   \\
10 & 1.5  &  74.4  & 87.8 & 72.2 & 30.8 & 35.2 & 70.2 & 93.4 & \textbf{78.3} & \textbf{19.7} & 24.4   \\
5 & 1.5  &  73.9  & 87.2 & 70.9 & 30.5 & 35.4 & \textbf{71.1} & 93.3 & 78.2 & 19.7 & \textbf{24.6}   \\
\midrule
30 & 2.1  & 72.5  & 87.9 & 71.1 & 28.1 & 33.1 & 65.8 & 92.5 & 74.6 & 14.9 & 18.6    \\
20 & 2.1  & 74.4  & 88.0 & 72.8 & 31.0 & 35.4 & 68.2 & 93.5 & 78.0 & 18.3 & 23.0    \\
15 & 2.1  & 75.1  & 88.0 & 72.6 & 31.0 & 35.4 & 68.7 & \textbf{93.5} & 78.2 & 19.5 & 23.9    \\
10 & 2.1  & \textbf{76.0}  & \textbf{88.6} & \textbf{73.3} & \textbf{31.6} & 35.7 & 69.7 & 93.4 & 78.2 & 19.5 & 24.0    \\
5 & 2.1  & 75.3  & 87.5 & 71.4 & 31.0 & \textbf{35.9} & 70.1 & 93.3 & 78.2 & 19.2 & 23.9    \\
\bottomrule % <-- Bottomrule here
\end{tabular}%}
\end{center}
\vspace{-5mm}
\end{table*}

\renewcommand{\arraystretch}{2}
\begin{figure}
\captionsetup[subfigure]{labelformat=empty}
\centering
\resizebox{1.0\linewidth}{!}{
\begin{tabular}{c@{}c@{}c@{}c@{}c@{}c}
\tiny{$I$} & \tiny{$30$} & \tiny{$20$} & \tiny{$15$} & \tiny{$10$} & \tiny{$5$}  \\
\multirow{2}{*}{\includegraphics[width=\mywidth,  ,valign=m, keepaspectratio,] {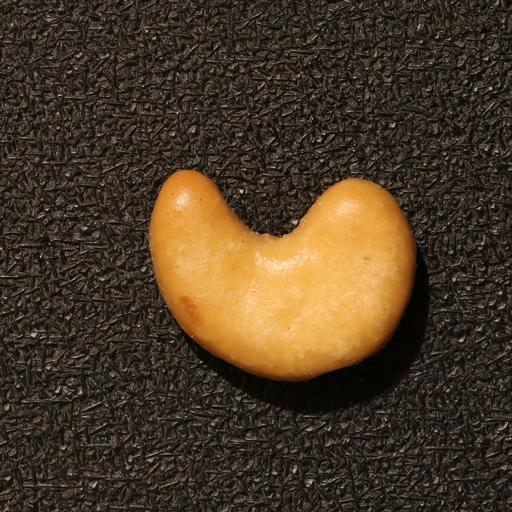}} & 
\includegraphics[width=\mywidth,  ,valign=m, keepaspectratio,] {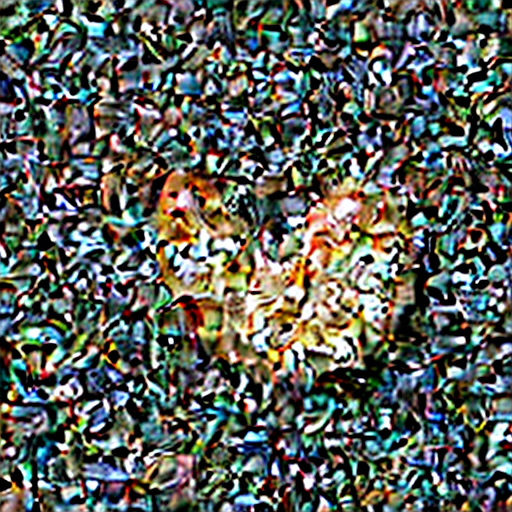} &
\includegraphics[width=\mywidth,  ,valign=m, keepaspectratio,] {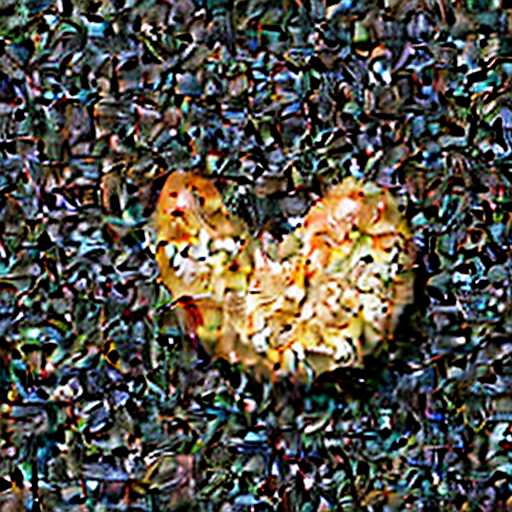} &
\includegraphics[width=\mywidth,  ,valign=m, keepaspectratio,] {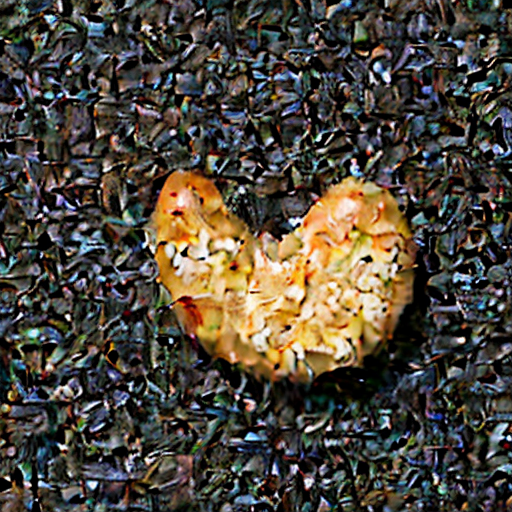} &
\includegraphics[width=\mywidth,  ,valign=m, keepaspectratio,] {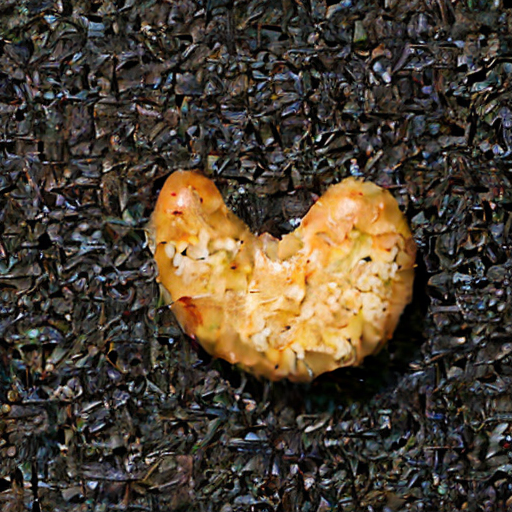} &
\includegraphics[width=\mywidth,  ,valign=m, keepaspectratio,] {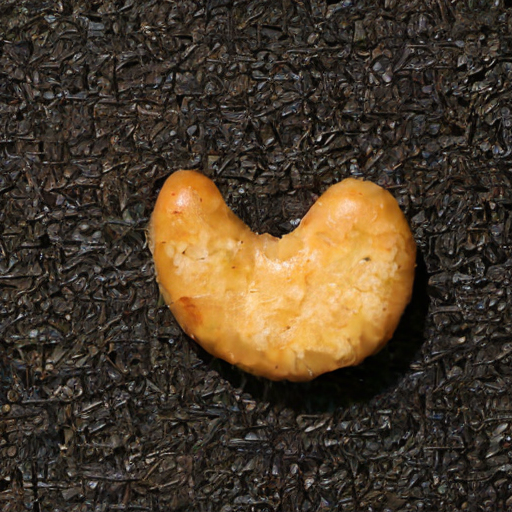}
\\ 
&  
\includegraphics[width=\mywidth,  ,valign=m, keepaspectratio,] {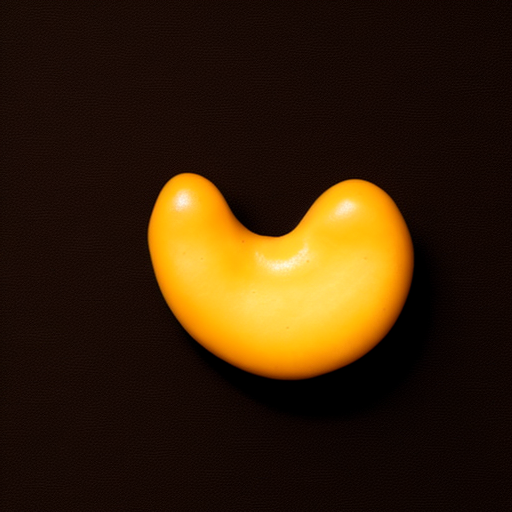} &
\includegraphics[width=\mywidth,  ,valign=m, keepaspectratio,] {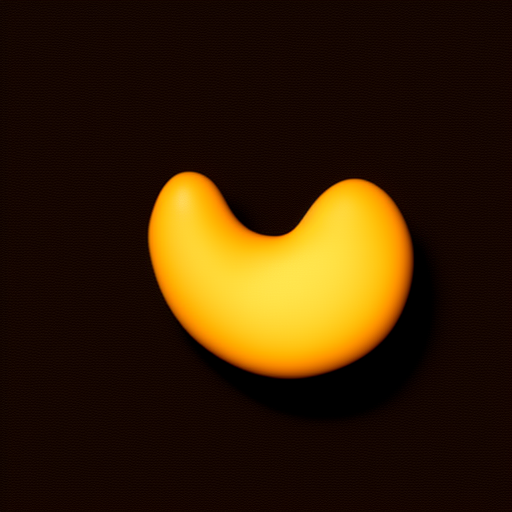} &
\includegraphics[width=\mywidth,  ,valign=m, keepaspectratio,] {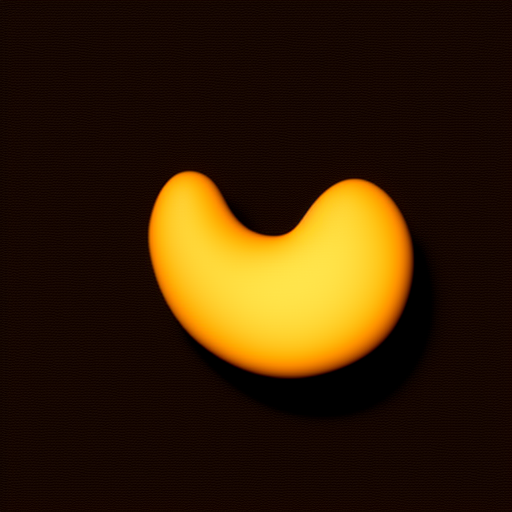} &
\includegraphics[width=\mywidth,  ,valign=m, keepaspectratio,] {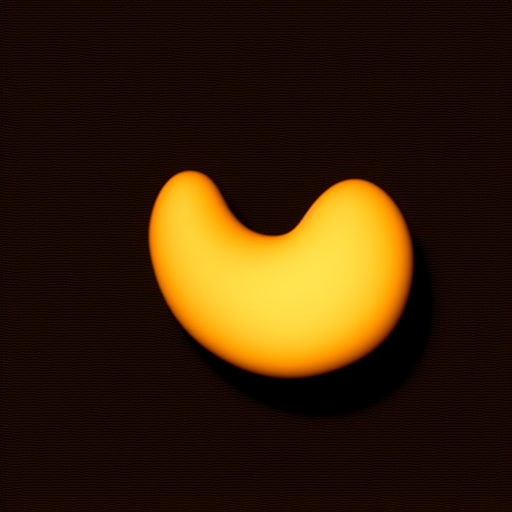} &
\includegraphics[width=\mywidth,  ,valign=m, keepaspectratio,] {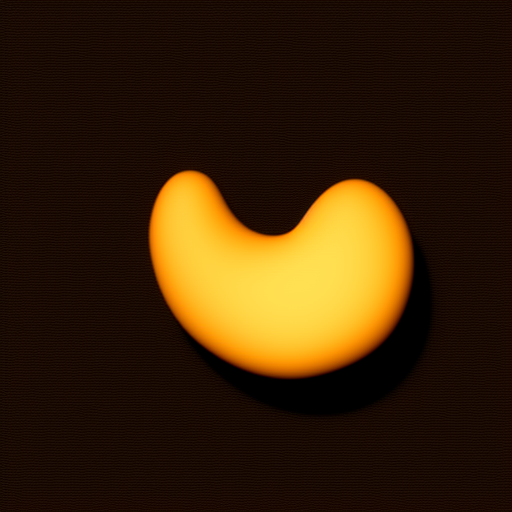} 
\\
\multirow{2}{*}{\includegraphics[width=\mywidth,  ,valign=m, keepaspectratio,] {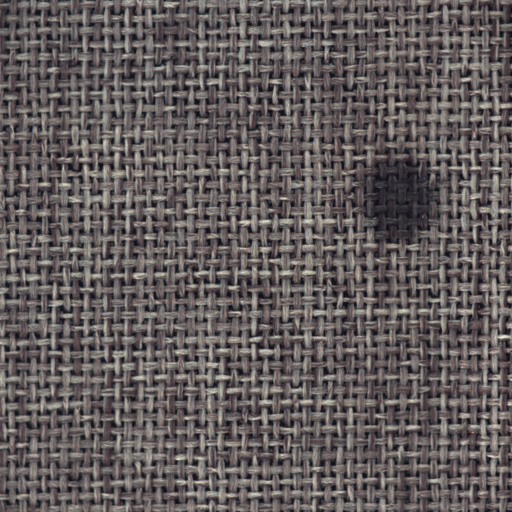}} & 
\includegraphics[width=\mywidth,  ,valign=m, keepaspectratio,] {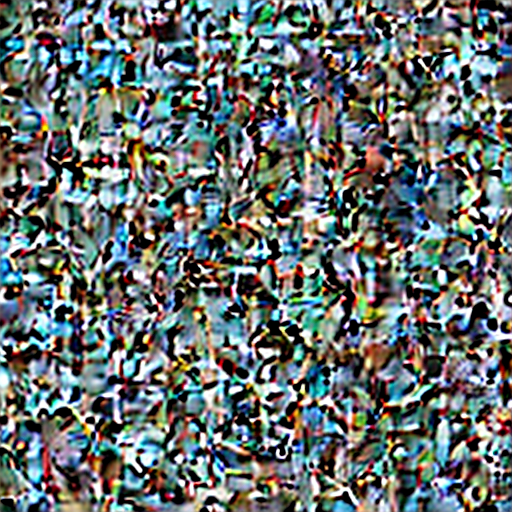} &
\includegraphics[width=\mywidth,  ,valign=m, keepaspectratio,] {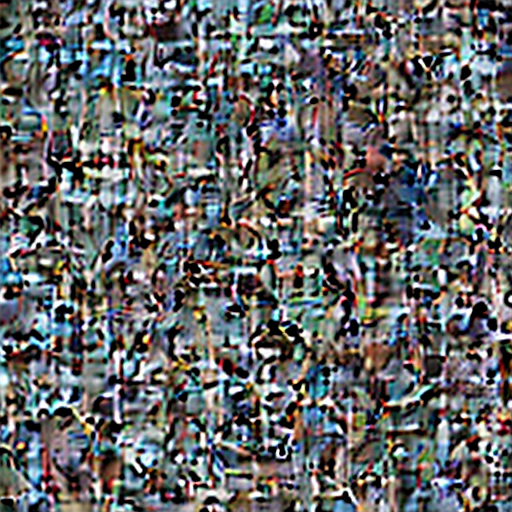} &
\includegraphics[width=\mywidth,  ,valign=m, keepaspectratio,] {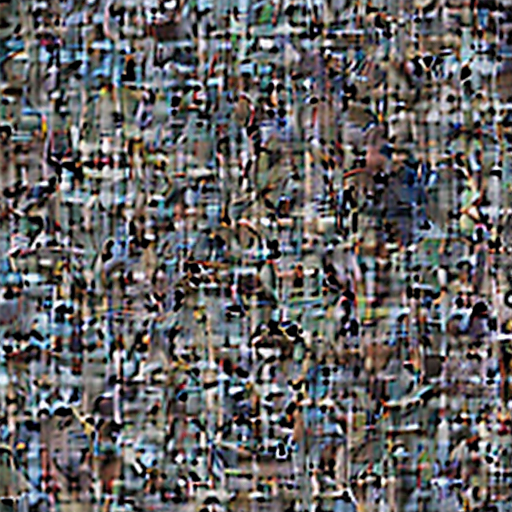} &
\includegraphics[width=\mywidth,  ,valign=m, keepaspectratio,] {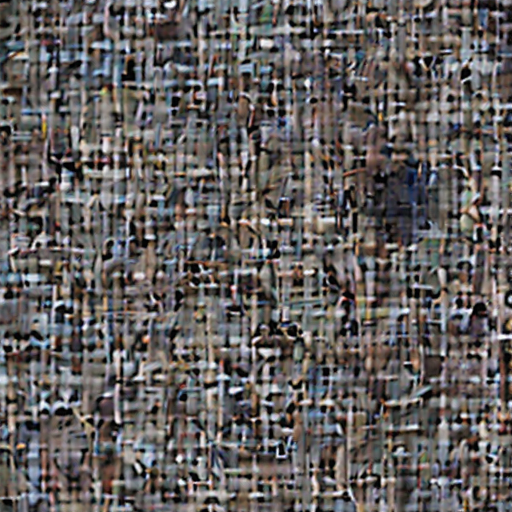} &
\includegraphics[width=\mywidth,  ,valign=m, keepaspectratio,] {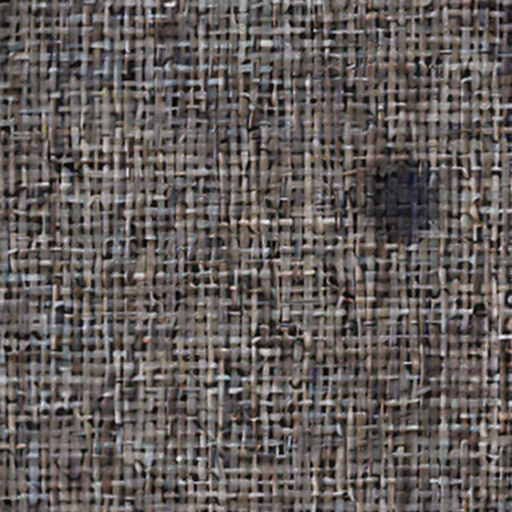}
\\ 
&  
\includegraphics[width=\mywidth,  ,valign=m, keepaspectratio,] {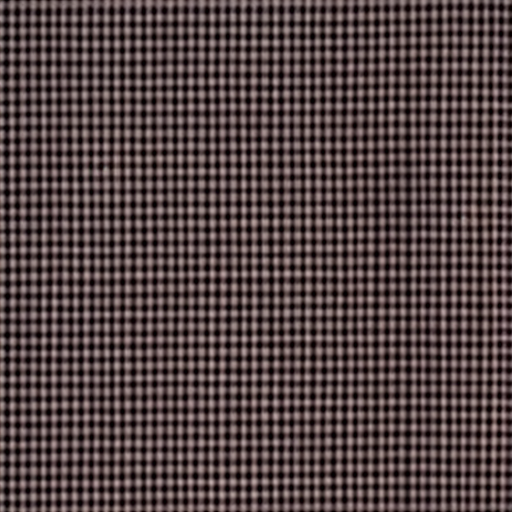} &
\includegraphics[width=\mywidth,  ,valign=m, keepaspectratio,] {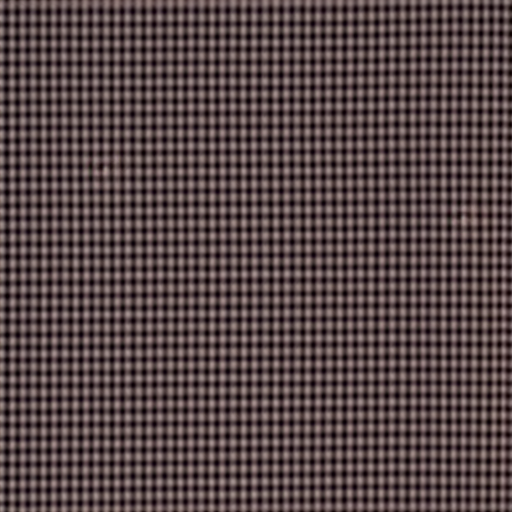} &
\includegraphics[width=\mywidth,  ,valign=m, keepaspectratio,] {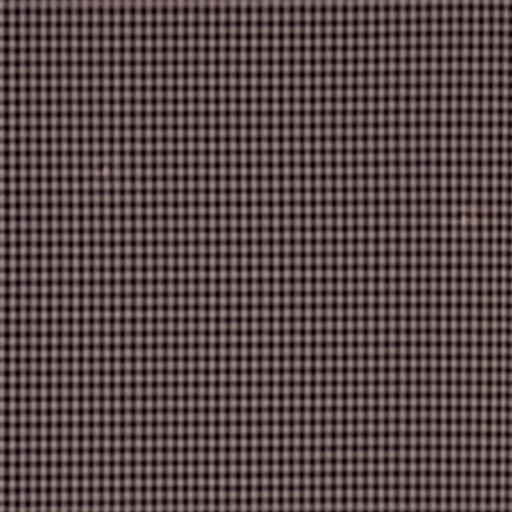} &
\includegraphics[width=\mywidth,  ,valign=m, keepaspectratio,] {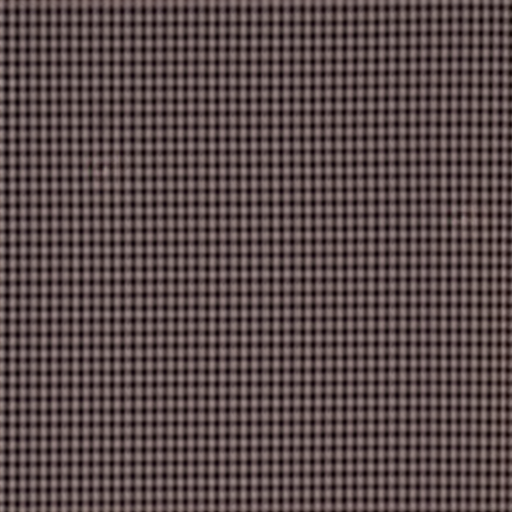} &
\includegraphics[width=\mywidth,  ,valign=m, keepaspectratio,] {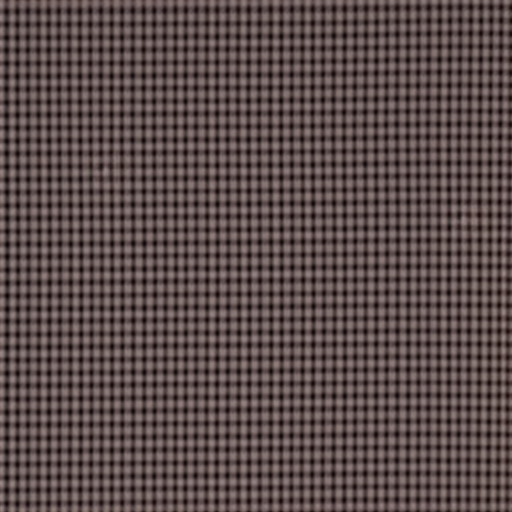} 
\end{tabular}}
\caption{Diffusion inversion visuals for an object on the top and a texture on the bottom. Starting from a small $T'$ captures more features from the input image as expected. A value between $10$ and $15$ achieves the best reconstruction.}
\label{fig:inversion_vis}
% \vspace{-7mm}
\end{figure}

On the MVTec-AD dataset, first of all, using the masks for the textures significantly reduces performance. This is expected since there is no dominant object; CutLER fails to segment the image. In our second experiment, we evaluate the impact of applying $M_{Obj}$ to object categories. Consistent with the texture results, the use of object masks leads to a performance drop across all metrics except $ROC_P$. This decline is primarily attributed to categories with large anomalous regions such as \textit{bottle} and \textit{metal nut} where the segmentation model struggles to produce accurate object masks, thereby negatively affecting performance. Notably, the PRO metric decreases by approximately $20$ points for both classes.
In order to have a unified framework for both datasets, we report final results by using $M_{Obj}$ on objects, but not on textures.

\renewcommand{\arraystretch}{1}
\begin{table}
\caption{Relative change with an empty prompt during inversion.}
\label{tab:abl_on_text}
\begin{center}
\resizebox{1.0\linewidth}{!}{
\begin{tabular}{c|ccccc}
 \toprule % <-- Toprule here
{Dataset} & $ROC_I$ & $ROC_P$ & $PRO$ & $AP_P$ & $F1_P$ \\
\midrule % <-- Midrule here
MVTec-AD  & -1.5   &  -6.7 & -14.6 & -11.2 & -9.3     \\
VISA   & -0.1 & -0.7 & -3.3 & -1.0 & -1.8    \\
\bottomrule % <-- Bottomrule here
\end{tabular}}
\end{center}
\vspace{-3mm}
\end{table}

\noindent\textbf{Effect of varying the timestep.} 
A critical hyperparameter in our method is to decide which timestep $T'$ should be used for denoising when the reverse sampling timestep $T$ is set $50$. In Table~\ref{tab:abl_on_timestep_sd}, we present the quantitative results for changing $T'$ from $30$ to $5$, i.e., reducing the added noise. See Figure~\ref{fig:inversion_vis} for examples of objects and textures. In both datasets, a $T'$ value between $5$ and $15$ gives the best overall results. This result shows the importance of \textbf{starting the denoising from a less contaminated latent vector.}

\noindent\textbf{Effect of changing SD version.} Different SD versions were examined, specifically $1.5$ and $2.1$. On MVTec-AD, SD $2.1$ achieved slightly better performance. Whereas on VISA, the best results are obtained with SD $1.5$. In order to have a consistent version, we chose SD $2.1$ and present the final results with it.

\noindent\textbf{Effect of using empty text input.} Our method only utilises the class name of the object as textual input. Although it requires no additional burden compared to deriving hundreds or even thousands of guided prompts, we ran experiments using an empty text as a condition to make it completely prompt-free. We present the relative change (c.f. Table~\ref{tab:mvtec_visa_scores}) when we omit the text input in Table~\ref{tab:abl_on_text}. On the VISA dataset, the difference is quite low, especially on the $ROC$ metrics. However, on the MVTEC-AD dataset, pixel-level metrics dropped significantly when no textual guidance is used, mainly due to the texture classes. 

%%%%%%%%%%%% Example Failure cases
% \begin{wrapfigure}{l}{0.5\textwidth}
\begin{figure}
\captionsetup[subfigure]{labelformat=empty}
\centering
\setlength\tabcolsep{1.5pt} % default value: 6pt
\resizebox{1.0\linewidth}{!}{
\begin{tabular}{c@{}c@{}c@{}c}

{\textit{\textbf{\tiny{Image}}}} & {\textit{\textbf{\tiny{Inv}}}} &  {\textit{\textbf{\tiny{Pred}}}} & {\textit{\textbf{\tiny{GT}}}}  \\

\includegraphics[width=\myw,  ,valign=m, keepaspectratio,] {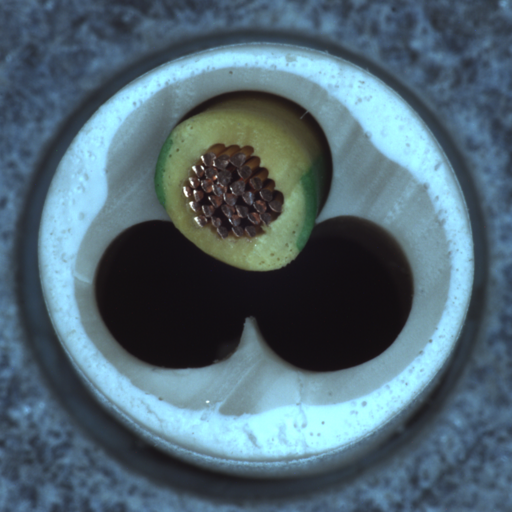} &
\includegraphics[width=\myw,  ,valign=m, keepaspectratio,] {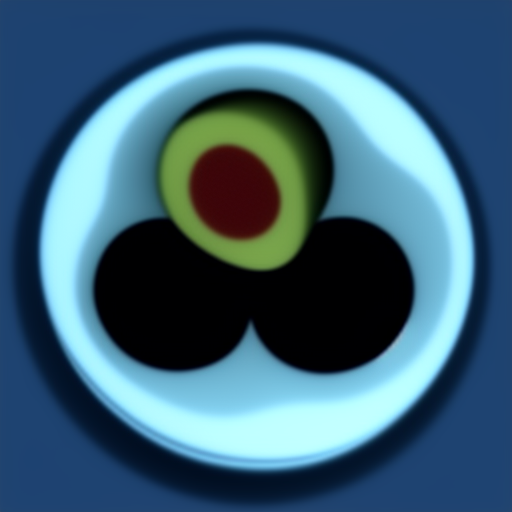}  &
\includegraphics[width=\myw,  ,valign=m, keepaspectratio,] {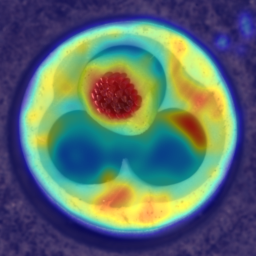}  &
\includegraphics[width=\myw,  ,valign=m, keepaspectratio,] {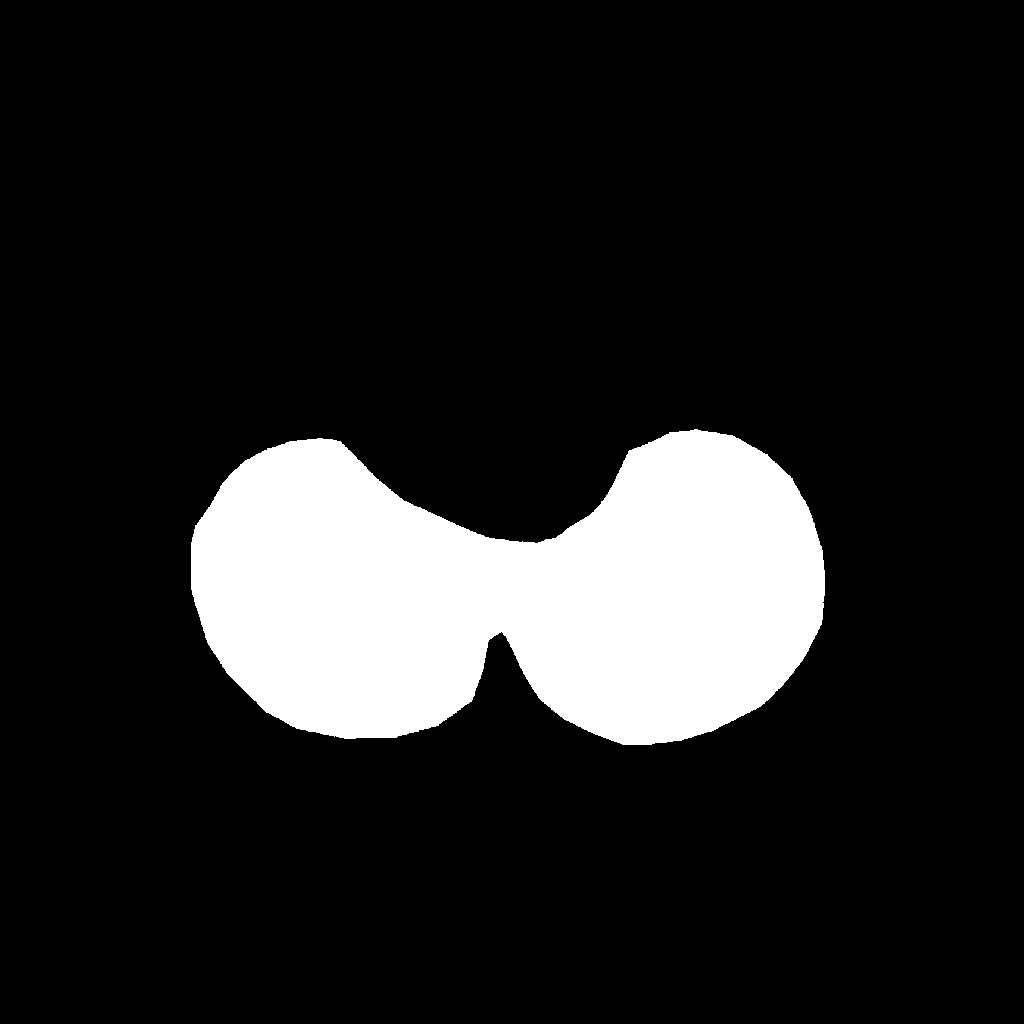}  \\

\includegraphics[width=\myw,  ,valign=m, keepaspectratio,] {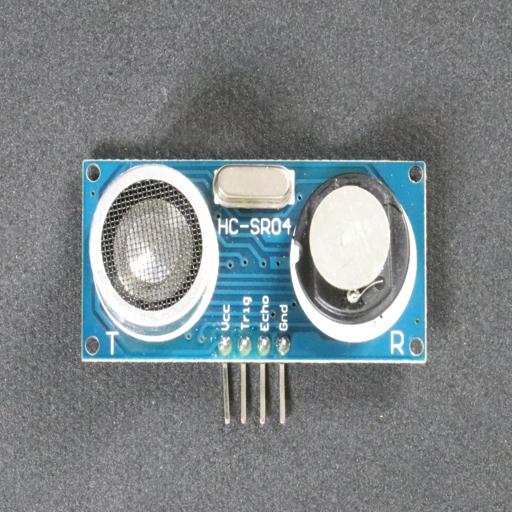} &
\includegraphics[width=\myw,  ,valign=m, keepaspectratio,] {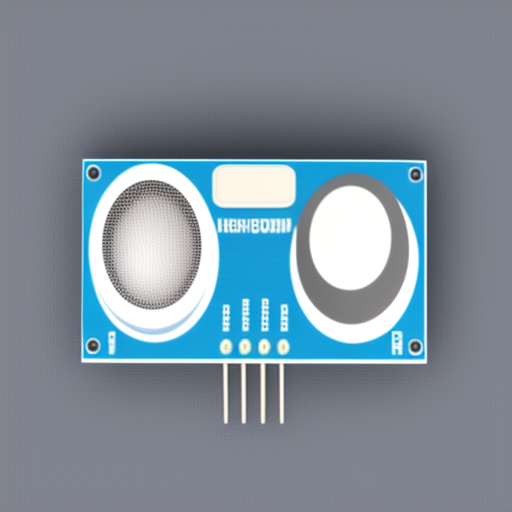}  &
\includegraphics[width=\myw,  ,valign=m, keepaspectratio,] {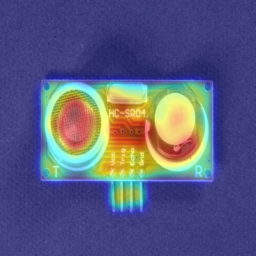} &
\includegraphics[width=\myw,  ,valign=m, keepaspectratio,] {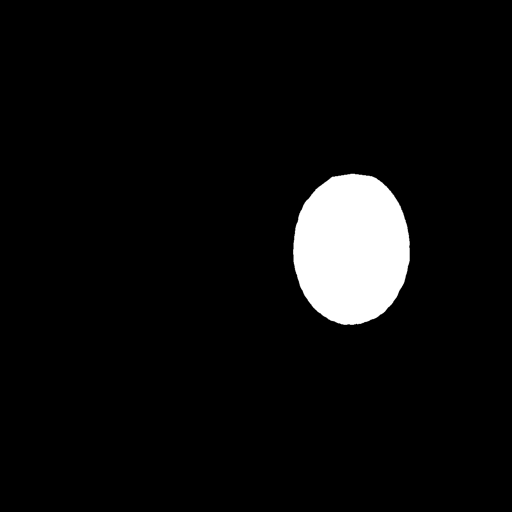} \\
\end{tabular}}
\caption{Example failure cases of {\ours}. Missing objects and contextual defects are not recoverable.}
\label{fig:failure_cases}
\vspace{-4mm}
\end{figure}

\subsection{Limitations}
\label{sec:limits}

Figure~\ref{fig:failure_cases} shows a couple of failure cases. For example, in the top row, when the anomaly is a missing part of the object, the inversion process could not complete the missing area. Another example, in the bottom row, when the object has many small details, such as soldering and pins, the inversion fails to reconstruct a correct image.  
Furthermore, we observe that our $ROC_I$ metric on MVTec-AD and VISA is lower than other baselines. This is due to the detection of anomalous pixels in images labelled as normal. These failures could be addressed by introducing a normal image example, which would extend our framework to few-shot based detection, a task we leave as future work.

\section{Conclusion}
\label{sec:conc}
Our method, {\ours}, highlights the potential of diffusion inversion as a powerful tool for ZSAD, offering a guided-prompt free, training-free, and vision-only alternative that achieves SoTA localisation performance on the VISA dataset. By eliminating the reliance on fine-grained prompts or auxiliary modalities, our approach simplifies the ZSAD pipeline while maintaining competitive performance. This work not only broadens the applicability of diffusion models in anomaly detection but also paves the way for future research into leveraging generative priors.
\\

\noindent\textbf{Acknowledgments.} This research was funded in whole or in part by the Luxembourg National Research Fund (FNR), grant reference \texttt{DEFENCE22/17813724/AUREA}.

{
    \small
    \bibliographystyle{ieeenat_fullname}
    \bibliography{main}
}

\clearpage
\setcounter{page}{1}

\section{Overview}

\begin{figure*}
\centering
\captionsetup[subfigure]{labelformat=empty}
\captionsetup{width=\textwidth}
\resizebox{1.0 \textwidth}{!}{
\begin{tabular}{l@{}c@{}c@{}c@{}c@{}c@{}c}
& \tiny{bracket black} & \tiny{bracket brown} & \tiny{bracket white} & \tiny{connector} & \tiny{metal plate} & \tiny{tubes} \\

{\rotatebox[origin=t]{90}{\textit{\tiny I}}}  & 
\includegraphics[width=\mywidth,  ,valign=m, keepaspectratio,] {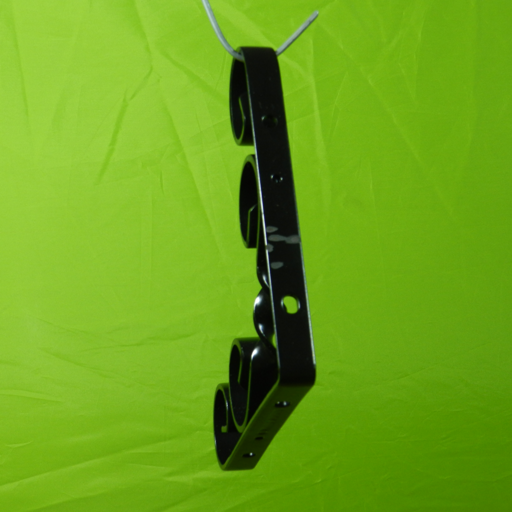} &
\includegraphics[width=\mywidth,  ,valign=m, keepaspectratio,] {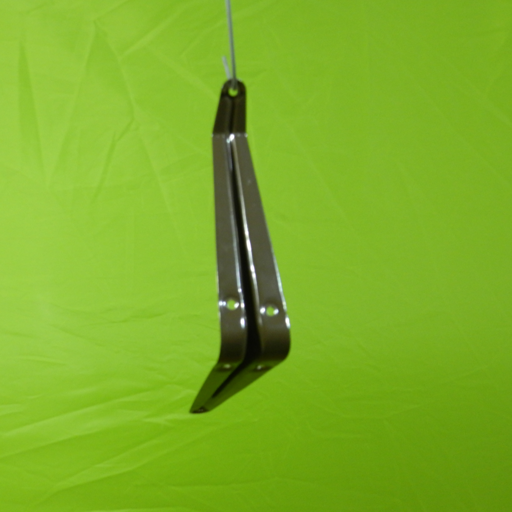} &
\includegraphics[width=\mywidth,  ,valign=m, keepaspectratio,] {./figures/our_results/mpdd/bracket_white/000.png} &
\includegraphics[width=\mywidth,  ,valign=m, keepaspectratio,] {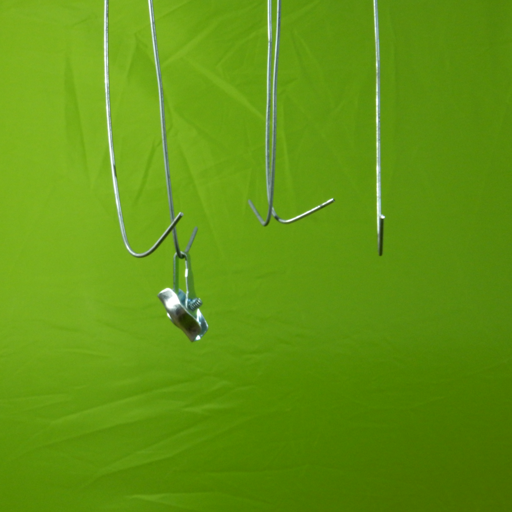} &
\includegraphics[width=\mywidth,  ,valign=m, keepaspectratio,] {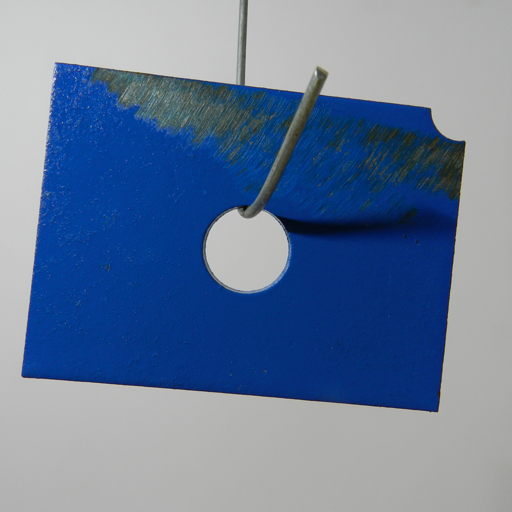} &
\includegraphics[width=\mywidth,  ,valign=m, keepaspectratio,] {./figures/our_results/mpdd/tubes/006.png}  
\\

{\rotatebox[origin=t]{90}{\textit{\textbf{ \tiny Inv.}}}}&  
\includegraphics[width=\mywidth,  ,valign=m, keepaspectratio,] {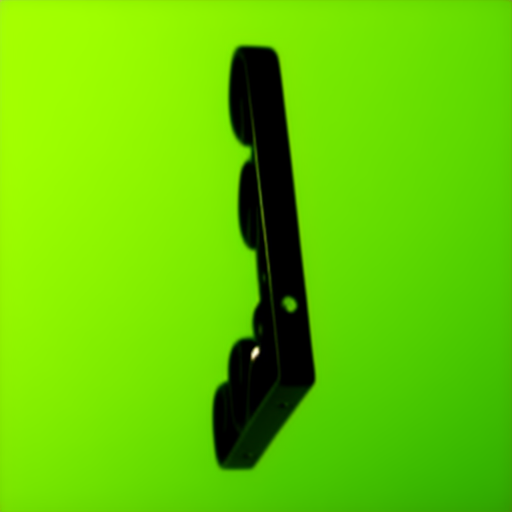} &
\includegraphics[width=\mywidth,  ,valign=m, keepaspectratio,] {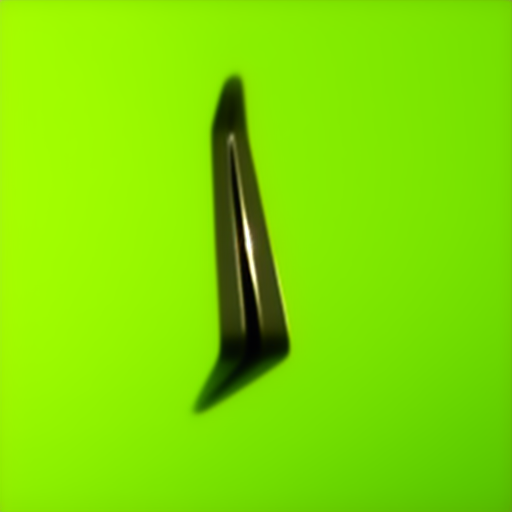} &
\includegraphics[width=\mywidth,  ,valign=m, keepaspectratio,] {./figures/our_results/mpdd/bracket_white/000.png_sampled.png} &
\includegraphics[width=\mywidth,  ,valign=m, keepaspectratio,] {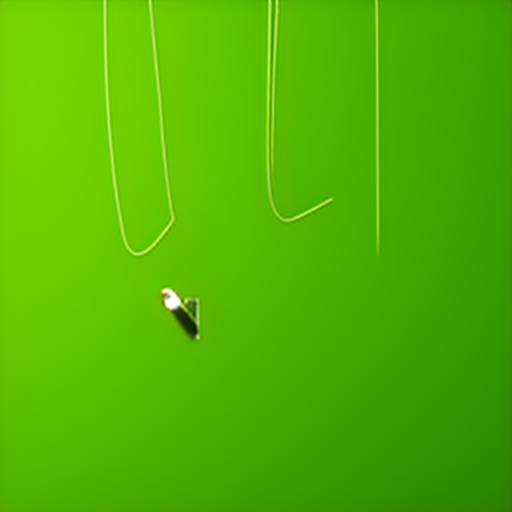} &
\includegraphics[width=\mywidth,  ,valign=m, keepaspectratio,] {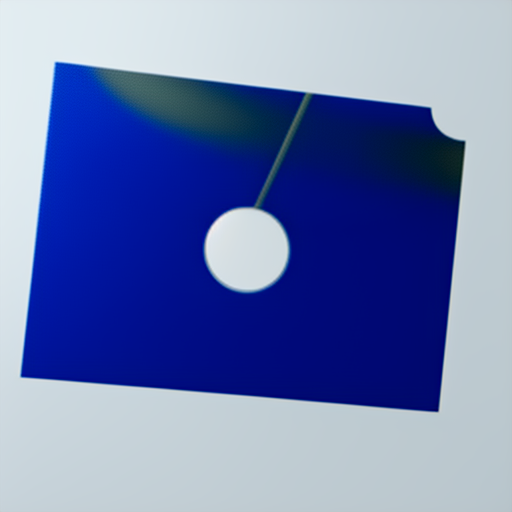} &
\includegraphics[width=\mywidth,  ,valign=m, keepaspectratio,] {./figures/our_results/mpdd/tubes/006.png_sampled.png} 
\\

{\rotatebox[origin=t]{90}{\textit{\textbf{\tiny {Pred.}}}}} & 
\includegraphics[width=\mywidth,  ,valign=m, keepaspectratio,] {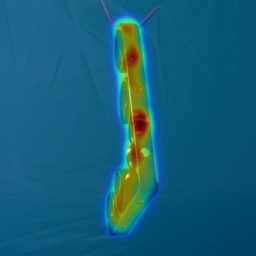} &
\includegraphics[width=\mywidth,  ,valign=m, keepaspectratio,] {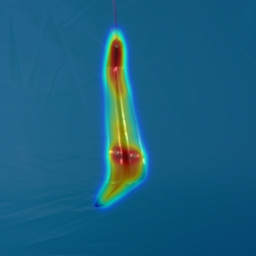} &
\includegraphics[width=\mywidth,  ,valign=m, keepaspectratio,] {./figures/our_results/mpdd/bracket_white/defective_painting_000.png-heatmap.png} &
\includegraphics[width=\mywidth,  ,valign=m, keepaspectratio,] {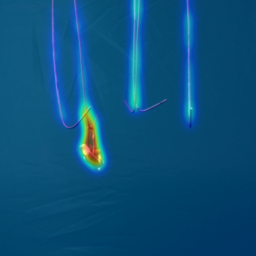} &
\includegraphics[width=\mywidth,  ,valign=m, keepaspectratio,] {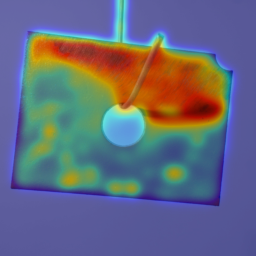} &
\includegraphics[width=\mywidth,  ,valign=m, keepaspectratio,] {./figures/our_results/mpdd/tubes/anomalous_006.png-heatmap.png} 
\\

{\rotatebox[origin=t]{90}{\textit{\textbf{\tiny GT}}}} & 
\includegraphics[width=\mywidth,  ,valign=m, keepaspectratio,] {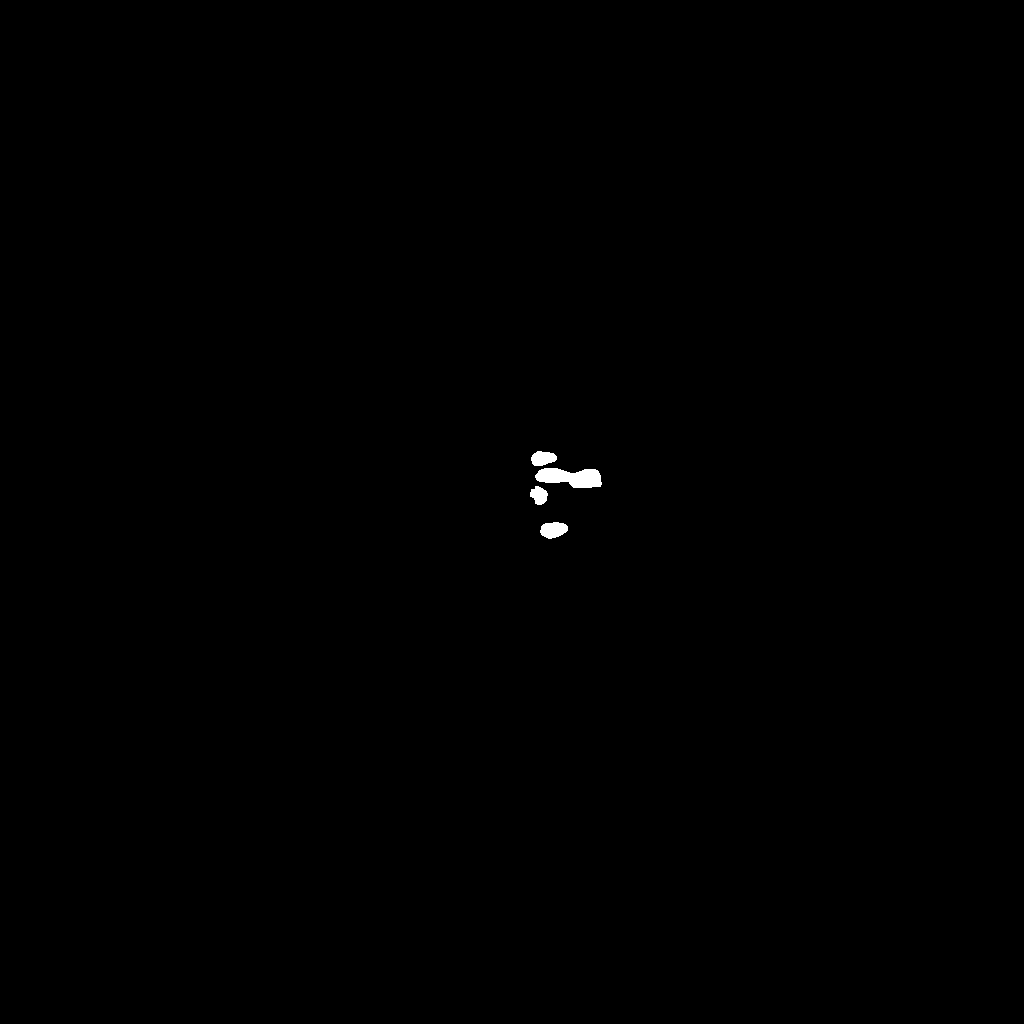} &
\includegraphics[width=\mywidth,  ,valign=m, keepaspectratio,] {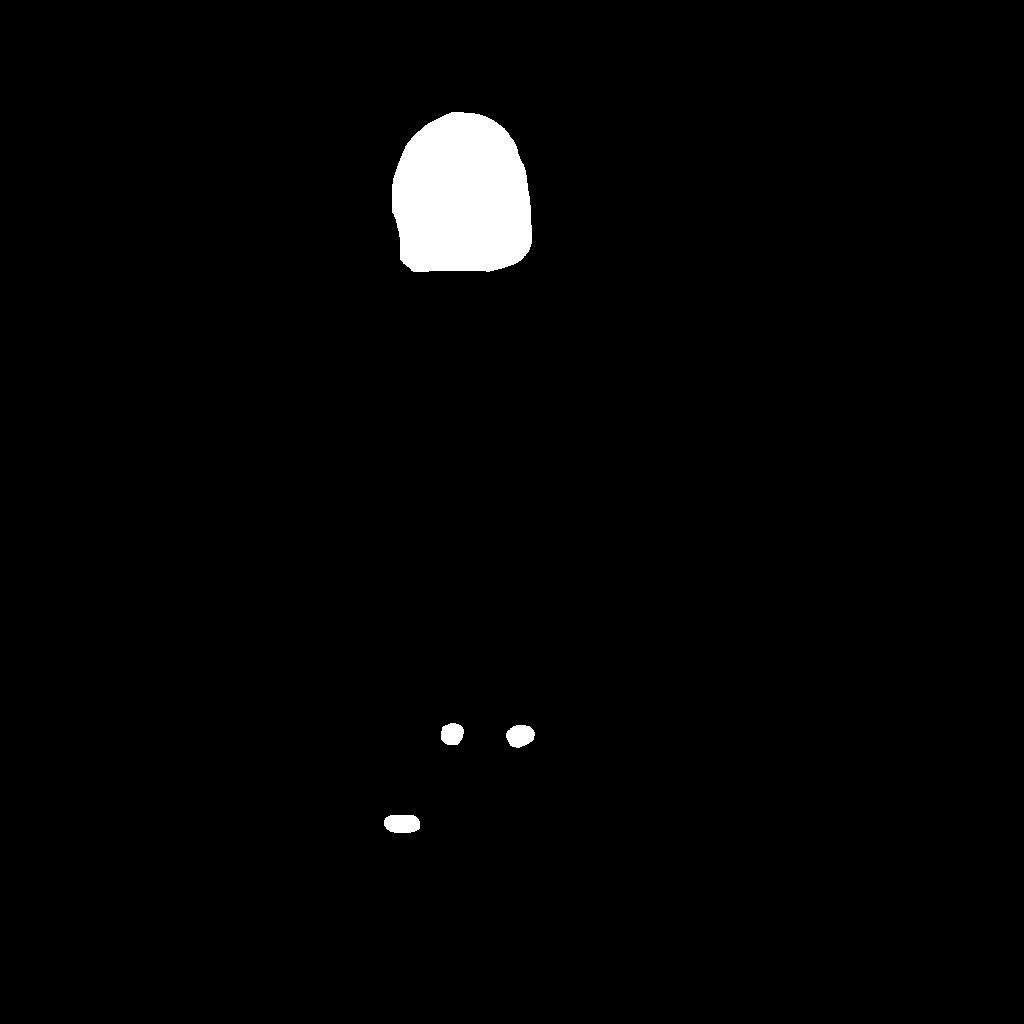} &
\includegraphics[width=\mywidth,  ,valign=m, keepaspectratio,] {./figures/our_results/mpdd/bracket_white/000_mask.png} &
\includegraphics[width=\mywidth,  ,valign=m, keepaspectratio,] {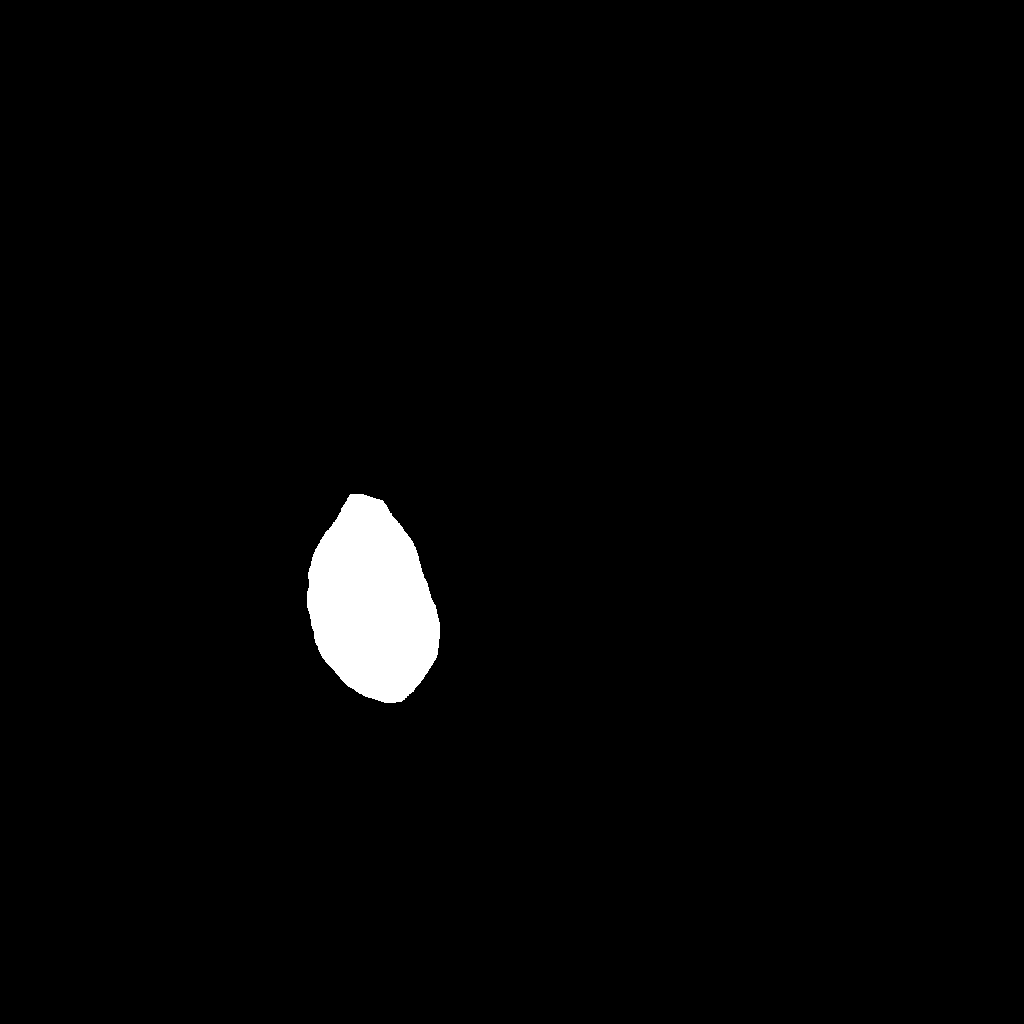} &
\includegraphics[width=\mywidth,  ,valign=m, keepaspectratio,] {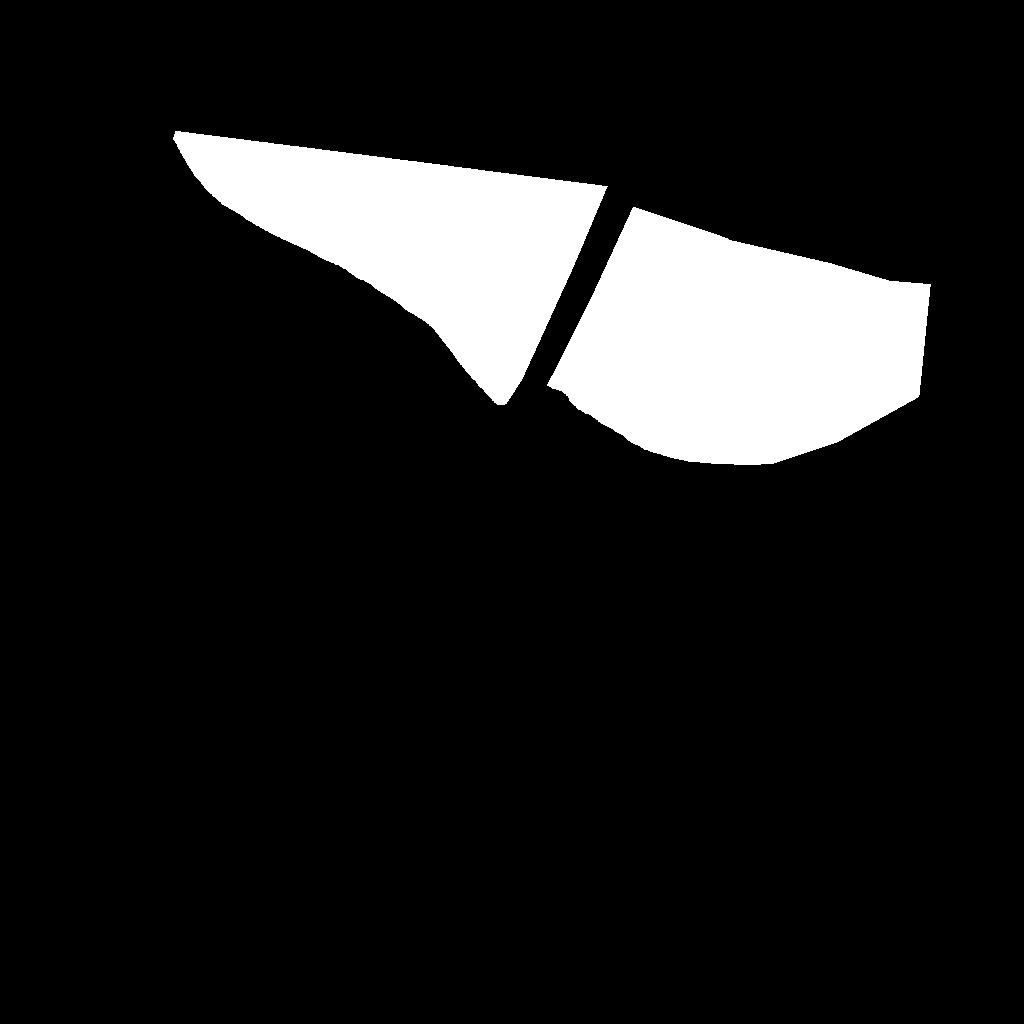} &
\includegraphics[width=\mywidth,  ,valign=m, keepaspectratio,] {./figures/our_results/mpdd/tubes/006_mask.png} 

\end{tabular}}
\captionof{figure}{Visual results of {\ours} on MPDD. {\ours} successfully detects the anomaly locations on various objects with different anomaly types.}
\label{fig:visual_results_mpdd_btad}
% \end{strip}
\end{figure*}

In this supplementary material, we first present qualitative results of {\ours} on the MPDD dataset, as shown in Figure~\ref{fig:visual_results_mpdd_btad}. We then provide detailed class-wise quantitative results across all three benchmark datasets.

On the MPDD dataset, {\ours} effectively localises various anomaly types, including \textit{misplaced holes}, \textit{defective paintings}, and \textit{rust} across diverse object categories. These results highlight the robustness and broad applicability of our approach in handling heterogeneous industrial anomalies.

In Table~\ref{tab:mvtec_scores_supp}, we provide a detailed class-wise comparison. We select two representative \textbf{training-free} ZSAD methods, FADE~\cite{zhu2024llms} and AnomalyVLM~\cite{cao2025personalizing}, for comparison, as (1) their official implementations are publicly available, and (2) they report state-of-the-art performance on at least one of the three benchmark datasets. To ensure a fair comparison, we reproduce their class-wise results using their official codebases. \textbf{Notably, we observed discrepancies between the results reported in the original papers and those obtained via their released implementations.} While we refer to the paper-reported numbers in the main text for consistency with prior work, here we report the values obtained through direct re-evaluation. \textbf{Importantly, these results show that our method achieves state-of-the-art performance across all pixel-level metrics, except for PRO, on the VISA dataset.}

\renewcommand{\arraystretch}{1}
\begin{table*}[ht]
\caption{Class-wise results on MVTec-AD (top), VISA (middle) and MPDD (bottom) datasets. }
\centering
\resizebox{1.0\textwidth}{!}{
\begin{tabular}{clccccc|ccccc|ccccc}
\toprule % <-- Toprule here
\multirow{3}{*}{Dataset} & \multirow{3}{*}{Class}   & \multicolumn{5}{c}{FADE~\cite{zhu2024llms}} & \multicolumn{5}{c}{AnomalyVLM~\cite{cao2025personalizing}} &  \multicolumn{5}{c}{{\ours}} \\ 
  \cmidrule(lr){3-7}  \cmidrule(lr){8-12}  \cmidrule(lr){13-17}
&  &  $ROC_{I}$ & $ROC_{P}$ &  {PRO} & $AP_P$ & $F1_P$ &  $ROC_{I}$ & $ROC_{P}$ &  {PRO} & $AP_P$ & $F1_P$ &  $ROC_{I}$ & $ROC_{P}$ &  {PRO} & $AP_P$ & $F1_P$\\
\midrule % <-- Midrule here
\multirow{16}{*}{\rotatebox[origin=c]{90}{MVTec-AD}}  
& bottle	& 97.9	& 89.2	& 79.6	& 48.9	& 50.6    & 79.3	& 69.0	& 39.5	& 32.6	& 39.3&     86.4 &   85.1 & 63.3 &  42.5	& 44.3	  \\
 & cable	& 85.1	& 78.4	& 68.0	& 13.0	& 20.9    & 55.0	& 69.1	& 33.6	& 19.9	& 32.3&     62.4 &   75.4 & 54.7 &  9.1	& 18.1	  \\
 & capsule	 & 68.9	& 89.2	& 86.3	& 14.8	& 21.5    & 52.0	& 61.9	& 34.8	& 8.3	& 17.7&     57.8 &   92.7 & 81.7 &  11.9	& 19.2	  \\
 & hazelnut & 91.4	& 97.3	& 92.0	& 44.1	& 47.5    & 83.9	& 85.5	& 59.9	& 38.8	& 44.8&     57.1 &   87.6 & 66.6 &  13.6	& 18.4	  \\
 & metalnut  & 95.9	& 70.8	& 69.7	& 21.2	& 30.5    & 28.9	& 60.0	& 23.9	& 13.3	& 36.2&     61.7 &   72.9 & 50.4 &  23.7	& 36.0	  \\
 & pill	   & 81.1	& 84.8	& 84.8	& 15.5	& 21.3    & 40.4	& 93.2	& 64.6	& 51.7	& 45.6&     61.9 &   82.4 & 65.1 &  11.2	& 18.4	  \\
 & screw	& 72.4	& 97.8	& 91.2	& 17.9	& 26.0    & 40.6	& 67.6	& 28.5	& 6.8	& 20.8&     66.1 &   95.8 & 75.9 &  6.1	& 12.3	  \\
 & toothbrush & 84.7& 90.7	& 89.3	& 15.6	& 23.4    & 20.6	& 64.2	& 25.3	& 2.2	& 8.2&      51.9 &   90.2 & 77.6 &  13.9	& 22.1	  \\
 & transistor & 87.5& 61.4	& 55.3	& 11.8	& 19.5    & 31.0	& 64.9	& 26.3	& 6.7	& 19.6&     70.4 &   71.6 & 38.8 &  9.2	& 17.8	  \\
 & zipper	& 89.7	& 92.4	& 82.9	& 37.4	& 42.6    & 35.2	& 78.8	& 28.4	& 12.4	& 21.3&     94.5 &   91.9 & 74.7 &  25.7	& 27.2	  \\
 & carpet	& 99.3	& 99.3	& 97.8	& 75.1	& 72.2    & 94.8	& 70.2	& 37.1	& 39.6	& 56.4&      99.8 &   98.7 & 96.2 & 76.0	& 71.1	  \\
 & grid	    & 99.5	& 98.3	& 94.7	& 38.4	& 47.2    & 80.5	& 63.9	& 42.7	& 12.4	& 18.2&      100.0 &   99.3 & 97.5 & 60.8	& 59.1  \\
 & leather	& 100.0	& 99.3	& 98.2	& 45.7	& 52.3    & 98.2	& 85.2	& 74.8	& 64.6	& 72.7&      97.5 &   99.0 & 97.9 & 54.3	& 57.2	  \\
 & tile	   & 99.9	& 92.3	& 79.5	& 47.0	& 53.6    & 96.3	& 81.6	& 63.5	& 61.7	& 64.8&      89.7 &   92.6 & 76.9 & 57.0	& 56.0	  \\
 & wood	   & 97.5	& 97.2	& 91.6	& 65.0	& 67.7    & 98.8	& 81.8	& 53.9	& 60.3	& 67.2&      82.9 &   93.5 & 83.0 & 59.0	& 58.3	  \\
 \cmidrule(lr){2-17}
 & mean	   & 90.1	& 89.2	& 84.1	& 34.1	& 39.8    & 62.4	& 73.1	& 42.4	& 28.8	& 37.7&      76.0 &  88.6 & 73.3 & 31.6	& 35.7	  \\
\midrule 

\multirow{13}{*}{\rotatebox[origin=c]{90}{VISA}}  
&candle    & 95.1	& 93.3	& 93.8	& 5.8	& 14.1    & 62.5	& 55.8	& 25.6	& 4.1	& 15.2                 &  69.8  &  95.3 & 78.7  & 8.3    & 14.9   \\  
&capsules  & 80.4	& 80.5	& 42.5	& 4.2	& 11.0    & 62.4	& 85.1	& 52.2	& 31.9	& 41.6                 &  57.7  &  94.5 & 86.4  & 32.2   & 39.0   \\    
&cashew    & 91.1	& 86.1	& 86.9	& 7.2	& 11.5    & 83.3	& 54.9	& 61.5	& 7.1	& 13.4                &  80.4  &  94.9 & 89.4  & 25.2   & 28.3   \\ 
&chewinggum& 95.0	& 99.1	& 90.3	& 35.7	& 47.0    & 92.6	& 93.8	& 52.6	& 31.3	& 36.1                 &  87.0  &  97.7 & 77.5  & 65.4   & 63.8   \\  
&fryum     & 69.1	& 94.0	& 90.4	& 24.3	& 32.6    & 38.8	& 94.5	& 34,0	& 17.6	& 22.5                 &  73.3  &  94.5 & 82.4  & 36.5   & 38.1   \\ 
&macaroni1 & 78.9	& 98.3	& 80.2	& 8.6	& 18.0    & 82.4	& 81.0	& 27.7	& 19.2	& 26.1                 &  76.4  &  97.0 & 89.2  & 20.0   & 28.2   \\  
&macaroni2 & 66.7	& 97.0	& 74.1	& 5.0	& 15.7    & 74.7	& 75.9	& 37.3	& 28.4	& 35.4                 &  66.2  &  96.8 & 88.2  & 8.8    & 16.4   \\  
&pcb1      & 72.5	& 83.9	& 69.8	& 2.2	&  5.9    & 67.9	& 74.0	& 3.9	& 6.1	& 21.3                 &  68.5  &  88.4 & 56.3  & 2.9    & 5.6    \\   
&pcb2      & 44.4	& 86.7	& 60.3	& 1.0	&  2.6    & 69.0	& 82.9	& 48.7	& 1.9	& 8.9                 &  66.4  &  85.1 & 56.1  & 1.7    & 4.9    \\   
&pcb3      & 63.8	& 87.3	& 72.7	& 1.3	&  3.2    & 55.6	& 72.7	& 47.8	& 1.9	& 9.8                 &  54.3  &  89.1 & 71.1  & 3.1    & 8.6    \\   
&pcb4      & 75.9	& 89.2	& 69.6	& 3.7	&  8.2    & 70.3	& 66.3	& 14.2	& 2.8	& 10.9                 &  55.3  &  92.6 & 73.5  & 11.7   & 19.4   \\     
&pipe-fryum& 62.8	& 96.9	& 94.8	& 21.1	& 32.1    & 93.1	& 72.4	& 37.5	& 13.0	& 24.5                 &  81.4  &  95.0 & 90.0  & 18.0   & 21.2  \\
 \cmidrule(lr){2-17}
& mean     & 74.7	& 91.0	& 77.1	& 10.0	& 16.8    & 71.1	& 75.8	& 36.9	& 13.8	& 22.1                 &  69.7  &  93.4 & 78.2  & 19.5   & 24.0 \\

\midrule % <-- Midrule here 
\multirow{7}{*}{\rotatebox[origin=c]{90}{MPDD}} 
& bracket black   & 39.8	& 93.7	& 80.8	& 1.0	& 2.6  &   67.8 &	89.4 &	31.8 &	1.1	& 2.8                &  51.7  &  95.8  & 80.6   &  2.2  &  5.0\\
& bracket brown   & 35.5	& 94.3	& 90.3	& 7.2	& 14.0 &   26.7 &	65.1 &	22.0 &	1.8	& 7.4                 &  52.6  &  90.0  & 70.6   &  4.1  &  8.5\\
& bracket white   & 48.2	& 94.4	& 70.9	& 0.4	& 1.0 &    69.6 &	94.8 &	28.9 &	9.8	& 22.7              &  59.1  &  97.5  & 89.0   &  2.2  &  6.0\\
& connector       & 79.0	& 96.2	& 89.4	& 17.9 & 27.8 &   23.8 &	71.9 &	24.6 &	3.3	& 11.4               &  73.6  &  95.0  & 83.3   &  8.6  &  16.0\\
& metal plate     & 94.3	& 92.1	& 80.3	& 51.8 & 61.6 &   37.0 &	72.4 &	24.4 &	22.2& 	48.2               &  63.1  &  94.5  & 81.7   &  62.3  & 68.3 \\
& tubes           & 80.1	& 94.8	& 80.0	& 35.3 & 40.9 &   78.0 &	74.7 &	35.3 &	7.1	& 16.2               &  80.4  &  96.6  & 86.6   &  58.1  & 58.3 \\   
 \cmidrule(lr){2-17}
& mean            & 62.8	& 94.2	& 83.9	& 18.9 & 24.6 &   50.5 &	78.0 &	27.8 &	7.5	& 18.1               &  63.4  &  94.9  & 82.0   &  22.9  & 27.0 \\
\bottomrule % <-- Bottomrule here
\end{tabular}}
\label{tab:mvtec_scores_supp}
\end{table*}

\end{document}